\documentclass[10pt,journal,compsoc]{IEEEtran}
\ifCLASSOPTIONcompsoc
  \usepackage[nocompress]{cite}
\else
  \usepackage{cite}
\fi
\ifCLASSINFOpdf
\else
\fi

\usepackage{algorithm}
\usepackage{algorithmic}
\usepackage{booktabs}
\usepackage{amsmath}

\usepackage{listings}
\usepackage{hyperref}
\usepackage{mathtools}
\usepackage{multirow}
\usepackage{bbm}
\usepackage{amsfonts,amssymb}
\usepackage{color}
\usepackage{graphicx}
\usepackage{subfig} 
\usepackage{subfloat}
\usepackage{amsthm}

\begin{document}
\title{Towards Effective Clustered Federated Learning: A Peer-to-peer Framework with Adaptive Neighbor Matching}

\author{Zexi~Li,
        Jiaxun~Lu,~\IEEEmembership{Member,~IEEE,}
        Shuang~Luo,
        Didi~Zhu,
        Yunfeng~Shao,~\IEEEmembership{Member,~IEEE,}
        Yinchuan~Li,~\IEEEmembership{Member,~IEEE,}
        Zhimeng Zhang,
        Yongheng Wang,
        and~Chao~Wu% <-this % stops a space
\IEEEcompsocitemizethanks{\IEEEcompsocthanksitem Zexi Li and Didi Zhu are with the College of Computer Science and Technology, Zhejiang University, Hangzhou, China.
E-mail: \{zexi.li, didi\_zhu\}@zju.edu.cn.
\IEEEcompsocthanksitem Jiaxun Lu, Yunfeng Shao and Yinchuan Li are with Huawei Noah’s Ark Lab, Beijing, China. E-mail: \{lujiaxun, shaoyunfeng, liyinchuan\}@huawei.com.
\IEEEcompsocthanksitem Chao Wu and Shuang Luo are with the School of Public Affairs, Zhejiang University, Hangzhou, China. E-mail: \{luoshuang, chao.wu\}@zju.edu.cn.
\IEEEcompsocthanksitem Zhimeng Zhang is with the School of Software Technology, Zhejiang University, Hangzhou, China. E-mail: zhimeng@zju.edu.cn.
\IEEEcompsocthanksitem Yongheng Wang is with the Big Data Intelligence Research Center, Zhejiang Lab, Hangzhou, China. E-mail: wangyh@zhejianglab.com.
\IEEEcompsocthanksitem Corresponding Author: Chao Wu and Yongheng Wang.}}

\markboth{Li \MakeLowercase{\textit{et al.}}: Towards Effective Clustered Federated Learning: A Peer-to-peer Framework with Adaptive Neighbor Matching}{Shell \MakeLowercase{\textit{et al.}}: Bare Demo of IEEEtran.cls for Computer Society Journals}

\IEEEtitleabstractindextext{%
\begin{abstract}
In federated learning (FL), clients may have diverse objectives, and merging all clients' knowledge into one global model will cause negative transfer to local performance. Thus, clustered FL is proposed to group similar clients into clusters and maintain several global models. In the literature, centralized clustered FL algorithms require the assumption of the number of clusters and hence are not effective enough to explore the latent relationships among clients. In this paper, without assuming the number of clusters, we propose a peer-to-peer (P2P) FL algorithm named \texttt{PANM}. In \texttt{PANM}, clients communicate with peers to adaptively form an effective clustered topology. Specifically, we present two novel metrics for measuring client similarity and a two-stage neighbor matching algorithm based Monte Carlo method and Expectation Maximization under the Gaussian Mixture Model assumption. We have conducted theoretical analyses of \texttt{PANM} on the probability of neighbor estimation and the error gap to the clustered optimum. We have also implemented extensive experiments under both synthetic and real-world clustered heterogeneity. Theoretical analysis and empirical experiments show that the proposed algorithm is superior to the P2P FL counterparts, and it achieves better performance than the centralized cluster FL method. \texttt{PANM} is effective even under extremely low communication budgets.
\end{abstract}

}

\maketitle

\IEEEdisplaynontitleabstractindextext

\IEEEpeerreviewmaketitle

\IEEEraisesectionheading{\section{Introduction}\label{intro}}

\IEEEPARstart{T}he proliferation of smart devices such as mobile phones, cameras, and sensors has dramatically expanded the perception of edge intelligence, increasingly forming
an Internet of Things (IoT) network \cite{lin2017survey,imteaj2021survey,nguyen2021federated}. The massive data from edge devices is key in generating powerful predictive models to provide better services to users. However, transferring the edge data to the data server poses a high privacy risk and high communication traffic burden, which renders the traditional centralized training ineffective. Therefore, Federated Learning (FL) \cite{DBLP:journals/tist/YangLCT19,DBLP:conf/aistats/McMahanMRHA17,DBLP:journals/spm/LiSTS20,DBLP:journals/ftml/KairouzMABBBBCC21} is proposed to facilitate collaborative training among edge clients without transferring the data to the cloud server. FL guarantees user privacy, reduces communication latency, and enhances learning performance, and it has broad and promising applications in IoT systems \cite{imteaj2021survey,nguyen2021federated}. 

In the real-world practice of FL, heterogeneity is an inherent problem (i.e. the Non-IID problem) since clients may have heterogeneous data distributions and thus diverse optimization objectives (learning tasks) \cite{DBLP:journals/corr/abs-2106-06843}. Clustered heterogeneity is prevalent in users' data, which means that a small group of clients has similar data distributions while there is dominant inconsistency among different groups. It is very common in applications such as recommendation systems \cite{sarwar2002recommender,li2003clustering}. Therefore, clustered FL methods \cite{DBLP:conf/nips/GhoshCYR20,DBLP:journals/corr/abs-2005-01026,DBLP:journals/tnn/SattlerMS21,duan2021flexible} are proposed for better personalization by grouping clients into clusters and maintaining a global model in each cluster. The main challenge of clustered heterogeneity is that the latent similarity relationship among clients is unknown. Existing clustered FL researches adopt the conventional server-client communication pattern and estimate clients' cluster identities by iterative \cite{DBLP:conf/nips/GhoshCYR20} or hierarchical \cite{DBLP:journals/tnn/SattlerMS21} methods. These centralized clustered FL methods highly rely on the assumed number of clusters or the assumed hierarchical level. However, in the real-world environment, the clustered relationship is latent, and it is impossible to know the number of clusters as prior knowledge. Inappropriate estimations of clusters' numbers will cause bad convergence. Besides, other concerns, like the reliability and communication bandwidth issues \cite{DBLP:conf/nips/LianZZHZL17}, brought by the central server, also hinder the performance of these centralized clustered FL methods.

To address these challenges, in this paper, we transform the client clustering problem into a binary classification problem from a peer-to-peer (P2P) perspective. Under P2P communication, each client decides whether to accept an accessible client as its neighbor based on local similarity measurement. Clients will select the most similar peers as their neighbors to realize personalization. Once the neighbor estimation is established, a clustered communication topology will be inherently built without assuming the number of clusters. Our method is named as \textbf{P}ersonalized \textbf{A}daptive \textbf{N}eighbour \textbf{M}atching (\textbf{\texttt{PANM}}) and it is proved to be effective and robust in various clustered heterogeneity.

In addition to the advantages compared with centralized clustered FL, \texttt{PANM} has superior performance over P2P FL counterparts. In previous works, most P2P FL algorithms assume random or fixed communication topologies \cite{DBLP:journals/corr/abs-1901-11173,lalitha2018fully} and they focus on reaching the global consensus by optimization techniques \cite{DBLP:conf/icml/0004KSJ21,DBLP:conf/icml/00010KJS21}. But in clustered heterogeneity, the global consensus does not exist, and random or fixed communications will harm personalization and decrease the accuracy. On the contrary, our method achieves partial group consensus by realizing the adaptive topology.

Our main contributions are as follows.
\begin{itemize}
\item We propose two efficient, effective, and privacy-preserving metrics to evaluate the pair-wise similarity of client objectives in P2P FL. They are based on losses and gradients, respectively.
\item We present \texttt{PANM}, a novel clustered FL algorithm based on P2P communication. \texttt{PANM} enables clients to match neighbors with consistent objectives (same cluster identity), improving local performance.
\item We devise two stages in \texttt{PANM}, the first is neighbor selection based on Monte Carlo, and the second is neighbor augmentation based on Gaussian Mixture Model. We provide theoretical guarantees of \texttt{PANM}.
\item We conduct extensive experiments on different datasets, Non-IID degrees, and network settings under both synthetic and real-world clustered heterogeneity. It is shown that \texttt{PANM} outperforms all P2P baselines, including \texttt{Oracle} that has the prior knowledge of cluster identities. Compared with centralized clustered FL algorithms, \texttt{PANM} is more effective in exploring latent cluster structure and has better performance. 
\end{itemize}

The rest of this paper is organized as follows. Section \ref{sect:related} reviews related works in clustered FL and P2P FL. Section \ref{problem_formulation} provides basic formulation about P2P clustered FL. Section \ref{sect:method} elaborates the technical details of \texttt{PANM}, including the two metrics for measuring client similarity and the two-stage neighbor matching algorithm. Theoretical analysis is also included in Section \ref{sect:method}. Section \ref{sect:exp} presents the experimental results. Sections \ref{sect:discussion} and \ref{sect:conclusion} provide further discussions and the conclusion of this paper.

\section{Related Works} \label{sect:related}
\noindent\textbf{Clustered Federated Learning. }Clustered FL holds the Non-IID assumption that different groups of clients have their own optimization objectives, and it is usually used to realize better accuracy performance \cite{DBLP:conf/nips/GhoshCYR20,DBLP:journals/tnn/SattlerMS21} or better compression of model updates \cite{cui2022optimal,cui2021slashing}. In this paper, we focus on the effectiveness of clustered FL on test accuracy, especially local personalization. In personalized clustered FL \cite{DBLP:conf/nips/GhoshCYR20,DBLP:journals/tnn/SattlerMS21}, aggregating models in the same cluster will bring better personalization while aggregation among different clusters will cause negative transfer\footnote{In some works, it is found that generalized knowledge can be transferred among cluster centroids \cite{nguyen2022self,duan2021flexible}. However, this only happens when the clustered heterogeneity is not dominant. In this paper, we focus on more heterogeneous clustered FL, where negative transfer exists among different clusters.}. 

To group clients into clusters, the main challenge of clustered FL is measuring client similarity. There are mainly three types of measurement, based on losses\cite{DBLP:conf/nips/GhoshCYR20}, gradients \cite{DBLP:journals/tnn/SattlerMS21,fedgroup,duan2021flexible}, model weights \cite{DBLP:journals/tnn/SattlerMS21,nguyen2022self,DBLP:journals/corr/abs-2005-01026,DBLP:conf/ijcnn/BriggsFA20}, respectively. In loss-based measurement, clients receive several models and infer them on the local dataset, and the one with the smallest loss has maximal similarity \cite{DBLP:conf/nips/GhoshCYR20}. For model weight and gradient measurement, cosine distance \cite{DBLP:journals/tnn/SattlerMS21,DBLP:journals/corr/abs-2005-01026} or Euclidean distance \cite{DBLP:conf/ijcnn/BriggsFA20,nguyen2022self,DBLP:journals/corr/abs-2005-01026} are used in previous works. It is verified that clients with similar data distributions will have small Euclidean distances and large cosine similarities in gradients or model weights.

The methods for clustering in previous works can be divided into two streams, the K-means-based and the hierarchical. For the K-means-based approaches, \texttt{FedSEM} \cite{DBLP:journals/corr/abs-2005-01026} first implements the K-means method based on clients' Euclidean distances of model weights on the server to cluster clients. However, server-side K-means clustering is computationally expensive. To solve this issue, Duan \textit{et al.} \cite{fedgroup,duan2021flexible} use decomposed cosine similarity to speed up computation and design an efficient newcomer device cold start mechanism. Additionally, Ghosh \textit{et al.}\cite{DBLP:conf/nips/GhoshCYR20} propose an efficient algorithm \texttt{IFCA} by inherently applying K-means to the client side. \texttt{IFCA} keeps several global models, and clients iteratively choose which global model it is prone to contribute to based on local losses of global models. 
In another stream of works, hierarchical clustering methods are used to achieve better personalization. Sattler \textit{et al.} \cite{DBLP:journals/tnn/SattlerMS21} use a hierarchical optimal bi-partitioning algorithm based on cosine similarity of weights or gradients. By bi-partitioning, the method realizes a model tree from personalization to generalization. Further, Briggs \textit{et al.} \cite{DBLP:conf/ijcnn/BriggsFA20} design a hierarchical algorithm for a wider range of Non-IID settings,  and the method separates clusters of clients by the similarity of their local models to the global model. Additionally, Dem-AI is developed for building large-scale distributed and democratized machine learning systems \cite{nguyen2022self}, and it realizes a bottom-up hierarchical clustering with specialized–generalized duality.
%and makes both intra-cluster and inter-cluster aggregations 

Note that all the algorithms mentioned above in clustered FL rely on the assumption of the number of clusters (the K-means-based methods) or the level of hierarchy (the hierarchical techniques). However, the number of clusters is latent and cannot be obtained as prior knowledge, and if the hyperparameters are set inappropriately, the clustering performance will degrade.

\noindent\textbf{Peer-to-peer Federated Learning. }Peer-to-peer federated learning (P2P FL, also known as decentralized FL) alters the centralized topology of conventional FL, and it allows clients to communicate with limited neighbors \cite{warnat2021swarm,DBLP:journals/corr/abs-1901-11173}. There is a study comparing decentralized algorithms like gossip learning with centralized FL in terms of communication efficiency. It is found that the best gossip variants perform comparably to the best centralized FL algorithms overall \cite{DBLP:journals/jpdc/HegedusDJ21}. Early works related to P2P FL introduce the P2P FL problem under privacy constraints and provide theoretical guarantees. Lalitha \textit{et al.} \cite{DBLP:journals/corr/abs-1901-11173,lalitha2018fully} use a Bayesian-like approach to let clients collectively learn a model that best fits the observations over the entire network. Bellet \textit{et al.} \cite{DBLP:conf/aistats/BelletGTT18} make P2P FL differentially private and analyze the trade-off between utility and privacy. They mainly study P2P FL under the IID data assumption, but heterogeneity is prevalent in FL practices.

Recent works mostly discuss class imbalance heterogeneity and communication problems. First, to tackle class imbalance, Li \textit{et al.} \cite{DBLP:journals/corr/abs-2012-13063} use mutual knowledge distillation instead of weight averaging. Bellet \textit{et al.} \cite{DBLP:journals/corr/abs-2104-07365} elaborately design a topology from holistic perspective. However, without a central server, the holistic perspective is impractical, and it is hard for clients to form such topology with limited observations. Second, communication of P2P FL can be more efficient by sparsification \cite{DBLP:conf/icdcs/TangSC20}, adaptive partial gradient aggregation \cite{DBLP:journals/caaitrit/JiangH20}, and using max-plus linear system theory to compute throughput \cite{DBLP:conf/nips/MarfoqXNV20}. Most recently, swarm learning \cite{warnat2021swarm} has been brought up as a P2P FL customized for medical research, utilizing edge computing and blockchain as infrastructures, and it has attracted wide attention. It provides strong application practices of P2P FL. 

While we are formulating this paper, we find a related same-time work (\texttt{PENS}) that has the same motivation as ours but uses different methods \cite{DBLP:journals/corr/abs-2107-08517}. \texttt{PENS} adopts a two-stage strategy. In the first stage, clients select top $k$ peers as neighbors for aggregation from randomly sampled $l$ neighbor candidates in each round. After the first stage, clients select the peers that were selected from as neighbors more than ``\textit{the expected amount of times}'' in the first stage as \textit{permanent neighbors}. In the second stage, in each round, clients randomly choose $k$ neighbors for aggregation from \textit{permanent neighbors}. It is possible for \texttt{PENS} to have noisy neighbor estimations, and we analyze the superiority of \texttt{PANM} to \texttt{PENS} in Section \ref{sect:method} and \ref{sect:exp}.

\begin{table}[t] \footnotesize
\caption{Important notations in this paper.}
\centering
\begin{tabular}{c|l} 
\toprule
Notation& Meaning\\
\midrule
% \textit{i}& Client \textit{i}\\
\textit{n}& Number of all clients\\
\textit{r}& Number of clusters\\
\textit{a}& Number of clients within a cluster\\
\textit{d}& Number of data samples in a client\\
\textit{c}& Assumed number of clusters in \texttt{IFCA} \cite{DBLP:conf/nips/GhoshCYR20}\\
\textit{k}& Size of aggregation neighbor list\\
\textit{l}& Size of neighbor candidate list\\
$\tau$& Round interval of \texttt{NAEM} in the second stage\\
$\alpha$& Hyperparameter in the gradient-based metric\\
${\rm N}_{i}^{t}$& Neighbor list of client \textit{i} in round \textit{t}\\
${\rm B}_{i}^{t}$& Neighbor bag of client \textit{i} in round \textit{t}\\
${\rm C}_{i}^{t}$& Neighbor candidate list of client \textit{i} in round \textit{t}\\
${\rm S}_{i}^{t}$& Selected neighbors in EM-step \\
${\rm M}_{i}^{t}$& Union set of ${\rm C}_{i}^{t}$ and ${\rm S}_{i}^{t}$\\
${\rm H}_{i}^{t}$& Neighbor estimation list in EM-step of client \textit{i}\\
\bottomrule
\end{tabular}
\label{tab:notation}
\end{table}

\section{Problem Formulation} \label{problem_formulation}
We first set up the clustered heterogeneity following previous works \cite{DBLP:conf/nips/GhoshCYR20}. There are $r$ different data distributions (clusters), $\mathcal{D}^{1}, \dots, \mathcal{D}^{r}$, and that the $n$ clients are partitioned into $r$ disjoint clusters. It is assumed that every client $i\ ( i \in [n])$ which belongs to cluster $j\ ( j \in [r])$ contains IID data samples $\mathcal{D}_{i}$ drawn from $\mathcal{D}^{j}$. For simplicity, we assume every client has the same number of samples that $\forall i,j \in [n], |\mathcal{D}_{i}| = |\mathcal{D}_{j}| = d$.

We solve clustered FL by forming it into a personalized P2P FL problem, in which we learn the personalized models $\mathbf{w} = (\mathbf{w}_{1}, \dots, \mathbf{w}_{n})$ and the neighbor graph matrix $\textbf{G}$. The expression is borrowed from personalized decentralized joint learning \cite{lalitha2018fully}, and the key difference is that we set binary elements instead of continuous elements in $\textbf{G}$, because in clustered heterogeneity, the task is to form a neighbor graph where the same-cluster clients should be connected (set as 1) while the different-cluster should be disconnected (set as 0). The neighbor graph matrix $\textbf{G}$ is an $n \times n$ square matrix, $G_{i,j}$ refers to the $(i, j).$th entry of the matrix, and it indicates whether client $j$ is in the neighbor bag of client $i$, 1 for true and 0 for false. The diagonal elements $G_{i,i}$ are all set to 1. To optimize $\mathbf{w}$ and $\textbf{G}$, the joint optimization objective is
% \begin{small}
\begin{equation} \label{equa1}
\begin{split}
    \min_{\mathbf{w} \in (\mathbb{R}^{d})^{n} \atop \textbf{G} \in \mathbb{R}^{n \times n}}J(\mathbf{w}, \textbf{G}) &= \sum_{i=1}^{n}F_{i}(\mathbf{w}_{i}) 
    + \frac{\nu}{2}\sum_{i=1}^{n}\sum_{j=1}^{n}G_{i,j}\Vert\mathbf{w}_{i} - \mathbf{w}_{j} \Vert^{2}, \\
    &s.t. \quad g(\textbf{G}) = 0.
\end{split}
\end{equation}

There are two terms in this objective function. The first one is the sum of loss functions, each involving the personalized model and the local dataset. $F_{i}:\mathbb{R}^{d}\rightarrow\mathbb{R}$ is the loss function of client $i\ (i\in [n])$ on its local dataset, given by: $F_{i}(\mathbf{w}_{i})\coloneqq\mathbb{E}_{\xi_{i} \sim\mathcal{D}_{i}}\left[f(\mathbf{w}_{i},\xi_{i})\right]$, where $\xi_{i} \sim\mathcal{D}_{i}$ denotes a random sample drawn from the local dataset, $\mathbb{E}(\cdot)$ means expectation, and $f(\mathbf{w}_{i},\xi_{i})$ refers to the loss function of $\mathbf{w}_{i}$ on a sample $\xi_{i}$. The second term enables collaboration by encouraging two clients to have a similar model if they are in neighborhood relationship. The condition $g(\textbf{G}) = 0$ regularizes the graph matrix regarding the network topology. For instance, the matrix should have diagonal vector to be $\mathbf{1}$ and the Frobenius norm\footnote{The Frobenius norm of $\textbf{G}$ indicates the number of connections in the graph.} should be assigned to meet the network setup. Other functions that regularize the degree of each node, both in-degree and out-degree, can also be integrated in $g(\textbf{G})$. 
In this paper, we solve the objective in Equation \ref{equa1} by adaptively matching neighbors.

We then introduce the model update protocol in P2P FL. In P2P FL, each client $i$ first updates its local model by local training and receives models from the neighbors. Then it averages the neighbors' models together with its local model into a new one and starts a next-round training. The process of one communication round in P2P FL can be formulated into the following:
\begin{equation} \label{equa2}
\mathbf{w}_{i}^{t+1}=\mathbf{w}_{i}^{t}-\eta\nabla F_{i}\left(\mathbf{w}_{i}^{t}\right)+\sum_{j \in {\rm N}_{i}^{t}}\left(\mathbf{w}_{j}^{t}-\eta\nabla F_{j}\left(\mathbf{w}_{j}^{t}\right)\right).
\end{equation}
${\rm N}_{i}^{t}$ is the round $t$'s aggregation neighbors of client $i$, randomly sampled from the neighbor bag ${\rm B}_{i}^{t}, ({\rm N}_{i}^{t}\subseteq {\rm B}_{i}^{t}, |{\rm N}_{i}^{t}| = k, |{\rm B}_{i}^{t}| = m, k\leqslant m)$. The neighbor bag of client $i$ is defined as the set of indexes $j$ where $G_{i,j} = 1, i \neq j$, as ${\rm B}_{i}^{t} = \{ j: G_{i,j} = 1; j \in [n], j \neq i\}$. The process that aggregating with random sampled peers from the neighbor bag is known as random gossip communication \cite{DBLP:journals/corr/abs-2107-08517,DBLP:journals/jpdc/HegedusDJ21}. In previous P2P FL works, the neighbor bag for each client includes all the remaining peers, but in clustered heterogeneity, such communication is noisy. In \texttt{PANM}, we keep the neighbor bag small but pure in the first stage and augment the neighbor bag in the second stage.

To improve readability, we summarize our main notations mentioned before or soon later as in Table \ref{tab:notation}.

\begin{figure*}[t]
\centering
\includegraphics[width=2\columnwidth]{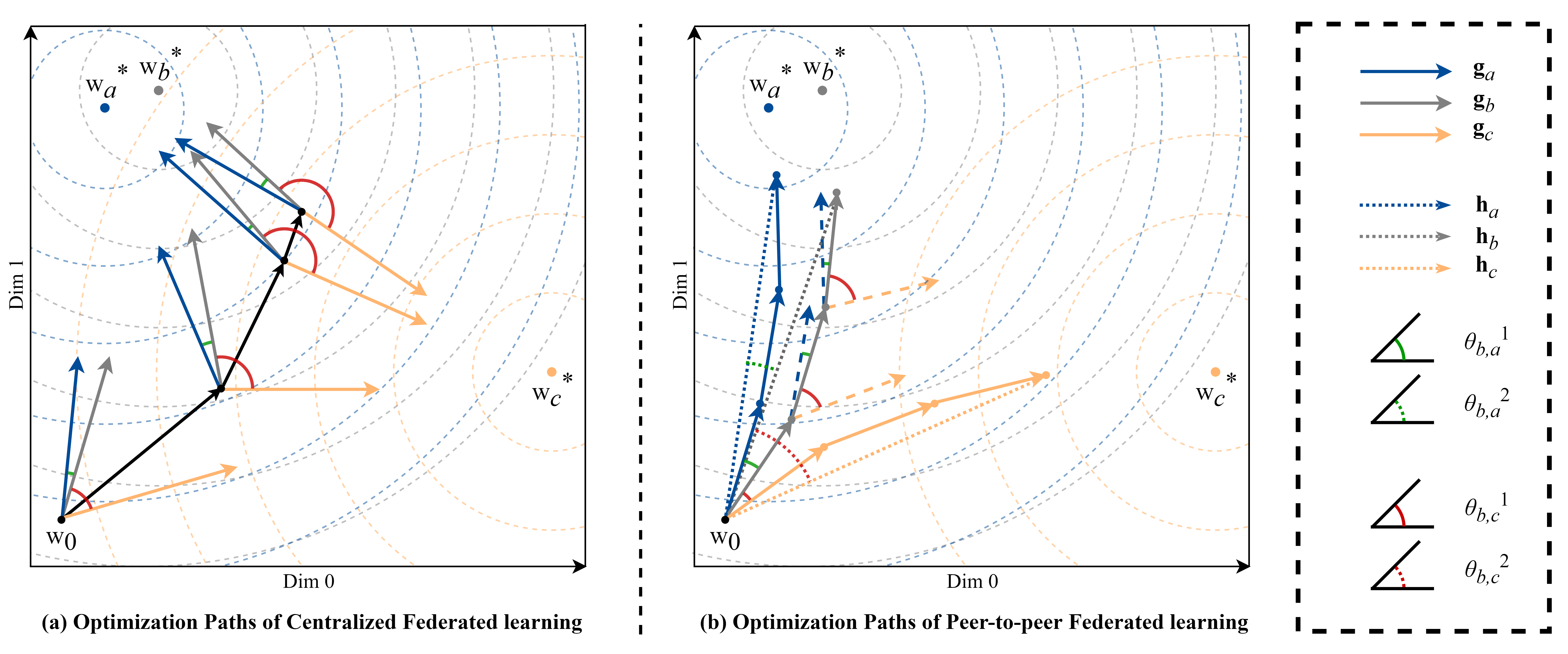}
\caption{Schematic diagram of optimization paths in centralized FL (a) and P2P FL (b), respectively. In the figure, objective of client \textit{a} is similar with client \textit{b} and dissimilar with client \textit{c}.}
\label{fig1}
\end{figure*}

\section{Method} \label{sect:method}
In this section, we will present two metrics for client similarity measurement in Section \ref{subsect:sim_metric}, neighbor selection based on Monte Carlo (first stage of \texttt{PANM}) in Section \ref{subsect:NSMC}, and neighbor augmentation based on EM-GMM in Section \ref{subsect:NAEM} (second stage of \texttt{PANM}). Then we will combine the similarity metrics, neighbor selection, and neighbor augmentation to devise \texttt{PANM} in Section \ref{subsect:panm}. Lastly, we will provide theoretical analysis in \ref{subsect:theory}.

\subsection{Metrics for Measuring Client Similarity} \label{subsect:sim_metric}
Metrics for measuring the consistency of optimization objectives are needed to enable clients to select same-cluster peers and filter out outliers. However, due to privacy concerns, in FL, we cannot use data distance measurements like maximum mean discrepancy distance \cite{tzeng2014deep}, since sharing data is forbidden. Loss evaluation is a simple metric, commonly used in the literature \cite{DBLP:conf/nips/GhoshCYR20,DBLP:journals/corr/abs-2107-08517}. In P2P FL, client $i$ receives client $j$'s model and infers the model on its local dataset. If the loss is small, it means client $j$ has a similar learning task with client $i$. We can use the reciprocal of loss as the similarity so that smaller loss indicates higher similarity, as
\begin{equation} \label{equ6}
{\rm s}_{i, j}=1/F_{i}(\mathbf{w}_{j}^{t} ),
\end{equation}
where ${\rm s}_{i,j}$ is the similarity between client $i$ and $j$ and $F_{i}(\mathbf{w}_{j}^{t})$ is the loss value of client $j$'s model validated on client $i$'s local dataset. Since this metric is simple, we adopt this metric in our \texttt{PANM}, named as \texttt{PANMLoss}. 

However, calculating loss value is computation-consuming because it requires inferring models on the training dataset. Moreover, local data may not be available for extra computation. Hence, we develop a more efficient metric based on gradients and accumulated weight updates.

In centralized clustered FL, Sattler \textit{et al.} \cite{DBLP:journals/tnn/SattlerMS21} use the cosine similarity of gradients to measure the consistency of optimization objectives, the function can be formulated as
\begin{equation} \label{equ3}
{\rm cos}~ \theta_{i,j}^{1}=\frac{\langle\textbf{g}_{i}^{t} , \textbf{g}_{j}^{t}\rangle}{\Vert\textbf{g}_{i}^{t}\Vert \cdot \Vert\textbf{g}_{j}^{t}\Vert}.
\end{equation}
In Equation \ref{equ3}, $\textbf{g}_{i}^{t}=\mathbf{w}_{i}^{t}-\mathbf{w}_{i}^{t-1}, \textbf{g}_{j}^{t}=\mathbf{w}_{j}^{t}-\mathbf{w}_{j}^{t-1}$. $\textbf{g}_{i}^{t}$ is the vectorized gradient of client $i$ in round $t$. $\textbf{g}_{i}^{t}$ and $\textbf{g}_{j}^{t}$ have $e$ dimensions, and $e$ is usually large in neural networks. ${\rm cos}~ \theta_{i,j}^{1}$ is the cosine function of the two gradient vectors, where $\theta_{i,j}^{1}$ refers to the angle of two vectors in the high-dimensional space. In centralized FL, models are initialized as the same global model at the beginning of local training in each round that $\mathbf{w}_{i}^{t-1}=\mathbf{w}_{j}^{t-1}=\mathbf{w}^{t-1}$, so the cosine function of gradients can effectively imply the consistency. We draw a 2-dimensional toy example to intuitively show the optimization trajectories and the angles of vectorized updates in Figure \ref{fig1}. The trajectory of centralized FL is shown in (a) of Figure \ref{fig1}. Although in P2P FL, client model weights diverge since the first round, the measurement of gradients will be noisy. The angle of gradients in P2P FL is shown as $\theta^{1}$ in (b) of Figure \ref{fig1}. 

To solve this issue, we notice the accumulated weight updates from the initial model can signify the history optimization directions, and the cosine similarity of the weight updates can also imply the consistency of objectives.
\begin{equation} \label{equ4}
{\rm cos}~ \theta_{i,j}^{2}=\frac{\langle\textbf{h}_{i}^{t} , \textbf{h}_{j}^{t}\rangle}{\Vert\textbf{h}_{i}^{t}\Vert \cdot \Vert\textbf{h}_{j}^{t}\Vert},
\end{equation}
where $\textbf{h}_{i}^{t} =  \mathbf{w}_{i}^{t}-\mathbf{w}_{0}, \textbf{h}_{j}^{t} = \mathbf{w}_{j}^{t}-\mathbf{w}_{0}$. $\textbf{h}_{i}^{t}$ is the vectorized accumulated weight updates of client $i$ from the initial model to the model in round $t$. Analogical to $\theta_{i,j}^{1}$, $\theta_{i,j}^{2}$ is the angle of two vectorized accumulated updates in the $e$-dimensional space, and we show $\theta^{2}$ in (b) of Figure \ref{fig1}.

According to Equations \ref{equ3} and \ref{equ4}, we combine the cosine functions of $\theta^{1}$ and $\theta^{2}$ to formulate our new metric as
\begin{equation} \label{equ5}
{\rm s}_{i,j}=\alpha \cos\theta_{i,j}^{1}+(1-\alpha)\cos\theta_{i,j}^{2},
\end{equation}
where $\alpha$ is the hyperparameter controlling the weight of two cosine functions, $\alpha \in [0, 1]$, we adopt the metric in Equation \ref{equ5} in \texttt{PANM}, notated as \texttt{PANMGrad}. We note that $\cos\theta_{i,j}^{1}$ has the same range as $\cos\theta_{i,j}^{2}$, which is $[-1, 1]$, so the outputs of two functions will have similar volumes. Thus, it is appropriate to set $\alpha$ around 0.5. Additionally, we notice that larger $\alpha$ will be beneficial when the P2P network is more densely connected. In a denser network, the clients' models are more synchronized, and it is more similar to the centralized FL, therefore, $\cos\theta_{i,j}^{1}$ will be more effective. Conversely, when the network is sparse, smaller $\alpha$ will help.

\begin{figure}[t]
\centering
\includegraphics[width=1\columnwidth]{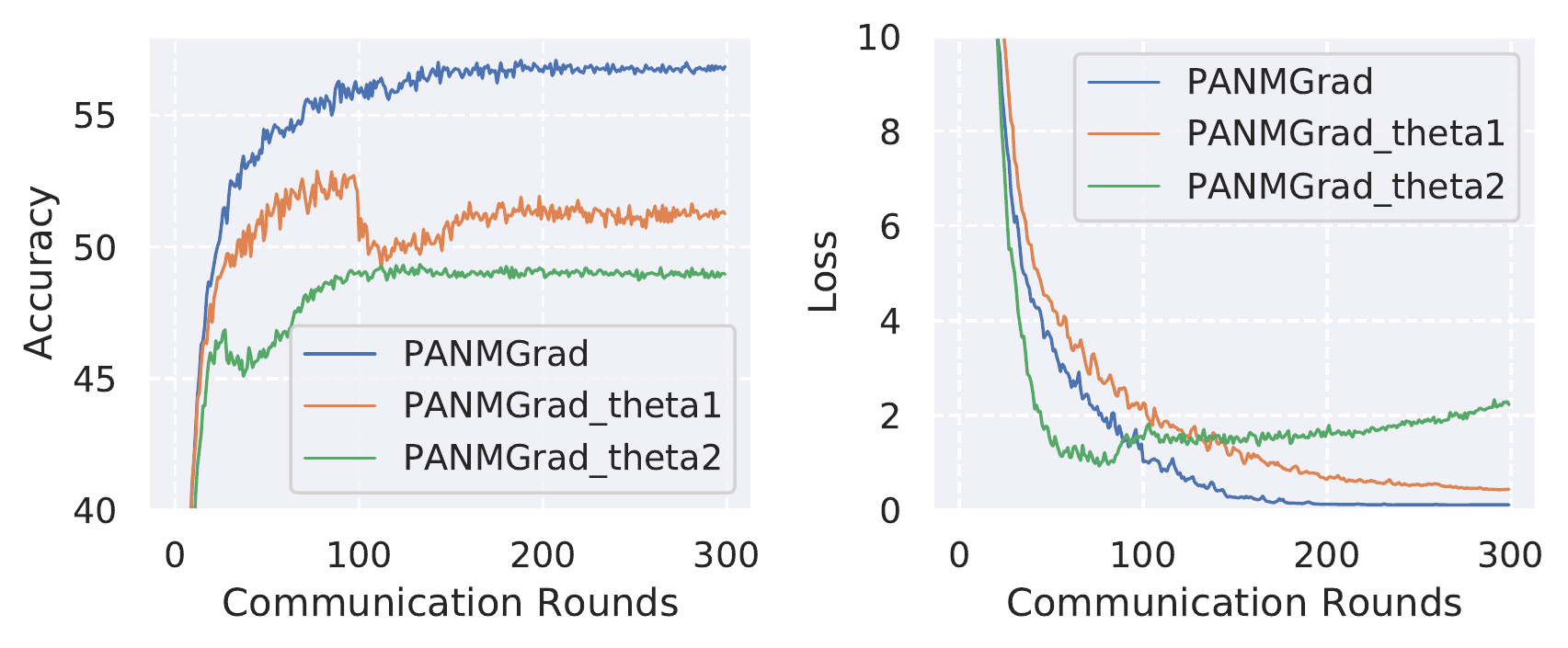} 
\caption{Ablation study of \texttt{PANMGrad}. CIFAR10 with two rotations \{0°,180°\}, 50 clients in each cluster, $l=10,k=5$, trainset size is 400. \texttt{PANMGrad} refers to \texttt{PANM} with metric in Equation \ref{equ5} ($\alpha=0.5$), \texttt{PANMGrad-theta1} refers to \texttt{PANM} with metric based on $\theta_{1}$, and \texttt{PANMGrad-theta2} refers to \texttt{PANM} with metric based on $\theta_{2}$.}
\label{ablationgrad}
\end{figure}

Notably, the combination of cosines $\theta_{1}$ and $\theta_{2}$ is more robust and effective compared with using one cosine function alone. We implement an ablation experiment as illustrated in Figure \ref{ablationgrad}. It is obvious that \texttt{PANMGrad} surpasses \texttt{PANMGrad-theta1} and \texttt{PANMGrad-theta2} by a large margin in accuracy curves. Besides, the curve of \texttt{PANMGrad} is stable and robust in both training stages (we will introduce the stages in Section \ref{subsect:NSMC} and \ref{subsect:NAEM}) while there are disturbances in the baseline curves. We explain the robustness and effectiveness of our metric by the complementarity of $\cos\theta^{1}$ and $\cos\theta^{2}$. $\cos\theta^{1}$ indicates the gradient direction of current round while $\cos\theta^{2}$ reflects historical neighbor-ship and optimization direction in previous rounds.

Moreover, we found the newly proposed gradient-based metric is computation-efficient compared with the loss-based metric. Without loss of generality, to simplify the analysis, we suppose the model is a $1 \times e$ vector, and a data sample is a $e \times 1$ vector, thus the computation of the inner product of gradients equals the computation of one data sample's inference. For client $i$, the number of local samples is $d$ and the number of neighbor candidates is $l$. To calculate the candidates' similarity, the loss-based metric has $\mathcal{O}(dl)$ computation complexity while the gradient-based metric has $\mathcal{O}(l)$ complexity. Usually, we have $d \gg l$, therefore the gradient-based metric is more efficient in computation. Although the loss-based metric can be more computation-efficient by reducing the dataset size, it will also reduce the effectiveness of measurement.

\subsection{Neighbor Selection Based on Monte Carlo} \label{subsect:NSMC}
\ 
\newline
Based on the similarity metrics mentioned in the last subsection, we can devise our P2P FL algorithm \texttt{PANM}. We introduce the first stage of \texttt{PANM} in this subsection.

In P2P FL, clients have access to receive models from randomly sampled peers (${\rm C}_{i}^{t}, |{\rm C}_{i}^{t}|=l$), and they need to select neighbors for model aggregation from these candidates. A natural way is to choose the top $k$ candidates with maximal similarities in each round, which is adopted in \texttt{PENS} \cite{DBLP:journals/corr/abs-2107-08517}. However, this method has constant expectation on the probability of correct neighbor estimation during the training process (we will theoretically prove it in Theorem \ref{thm1} and Corollary \ref{corollary1}). If $l,k$ are not appropriately set or the environment is more heterogeneous, the neighbor estimation will be constantly noisy. We solve this challenge by resorting to Monte Carlo method. We formulate the objective of the first stage as: \textit{for each client, find the most similar peers as its neighbors.} To reach this objective, the Monte Carlo method adds the neighbors in the last round together with the random sampled peers in the current round to the candidate list. As the number of rounds increases, the expected probability of the most similar neighbors increases, and if the similarity measurement is effective, these similar neighbors are the same-clustered peers. We name this method as Neighbor Selection Based on Monte Carlo (\texttt{\textbf{NSMC}}). We summarize both the first stage of \texttt{PENS} and \texttt{NSMC} in one equation, as
\begin{equation} \label{equ8}
\begin{split}
{\rm N}_{i}^{t}&= \mathop{\arg\max}_{{\rm N}}\sum\limits_{j \in {\rm N}} {\rm s}_{i, j}\\
&s.t.\ {\rm N}\subsetneqq {\rm C}_{i}^{t}\cup {\rm L}, |{\rm N}| = k.
\end{split}
\end{equation}
In Equation \ref{equ8}, for \texttt{PENS}, it has ${\rm L} = \emptyset$ in all rounds; and for CNI, it has ${\rm L} = \emptyset$ in the first round, and ${\rm L} = {\rm N}_{i}^{t-1}$ when $t > 1$. We will present theoretical analysis on \texttt{PENS} and \texttt{NSMC} in Section \ref{subsect:theory}. It is found that the expected probability of true neighbors (being the same-cluster) is rapidly increasing in \texttt{NSMC} and keeps constant in \texttt{PENS}. We also show the empirical results that \texttt{NSMC} can boost personalization even better than \texttt{Oracle} in Section \ref{sect:exp}. \texttt{NSMC} facilitates clients to select peers with globally maximal similarities and these most similar peers may be more helpful than other peers in the same cluster.

\subsection{Neighbor Augmentation Based on EM-GMM} \label{subsect:NAEM}
\ 
\newline
After \texttt{NSMC}, we enable clients to have few neighbors in the neighbor bag with high probability of being true neighbors, in other words, the precision is high. For clustered FL, the recall of clustering is also essential since each client needs to find out the whole community with the same objective. Thus, in the second stage of \texttt{PANM}, we facilitate clients to discover more peers with consistent objectives.

In the second stage of \texttt{PENS}, clients choose peers that are selected more than ``\textit{the expected amount of times}'' in the first stage as neighbors. But if the setting is difficult, stage-one neighbors of \texttt{PENS} are prone to be noisy, afterward, in the second stage, the matched neighbors are more likely to include outliers. Besides, \texttt{PENS} requires the hyperparameter ``\textit{the expected amount of times}'', and without prior knowledge of cluster information, it is hard to set the hyperparameter to an appropriate value. To better solve these problems, we propose a more effective neighbor bag augmentation method, which is based on Expectation Maximization of Gaussian Mixture Model (EM-GMM). 

For a client, given a set of randomly sampled peers, it is obvious that the true neighbors (i.e. with the same cluster identity) may have high similarities with it while the false ones (i.e. with different cluster identities) have low similarities. Therefore, we make the Gaussian Mixture Model assumption that the similarities of the true neighbors obey a consistent distribution while the similarities of the false ones obey another distribution. Formally, the assumption is shown in Assumption \ref{assump2}.
\newtheorem{assumption}{Assumption}
\begin{assumption} \label{assump2}
\textbf{(Gaussian Mixture Model Assumption)} For client $i$ ($\forall i \in [n]$), the similarities between the true neighbors and client $i$ obey a Gaussian distribution, parameterized by $\mathcal N(\mu_{0},\sigma_{0}^{2})$, and the similarities between the false neighbors and client $i$ obey another Gaussian $\mathcal N(\mu_{1},\sigma_{1}^{2})$ that
\begin{equation} 
\begin{split}
    {\rm s}_{i,p}\sim \mathcal N(\mu_{0},\sigma_{0}^{2}),&\ {\rm s}_{i,q}\sim \mathcal N(\mu_{1},\sigma_{1}^{2}) \\
    \forall p \in {\rm N}_{i}^*,&\ q \in \overline{{\rm N}_{i}^*},
\end{split}
\nonumber
\end{equation}
where ${\rm N}_{i}^*$ refers to the true neighbors of client $i$, and $\overline{{\rm N}_{i}^*}$ refers to the false neighbors of client $i$.
We have $\mu_{0} > \mu_{1}$.
\end{assumption}

\begin{figure}[t]
\centering
\includegraphics[width=0.37\columnwidth]{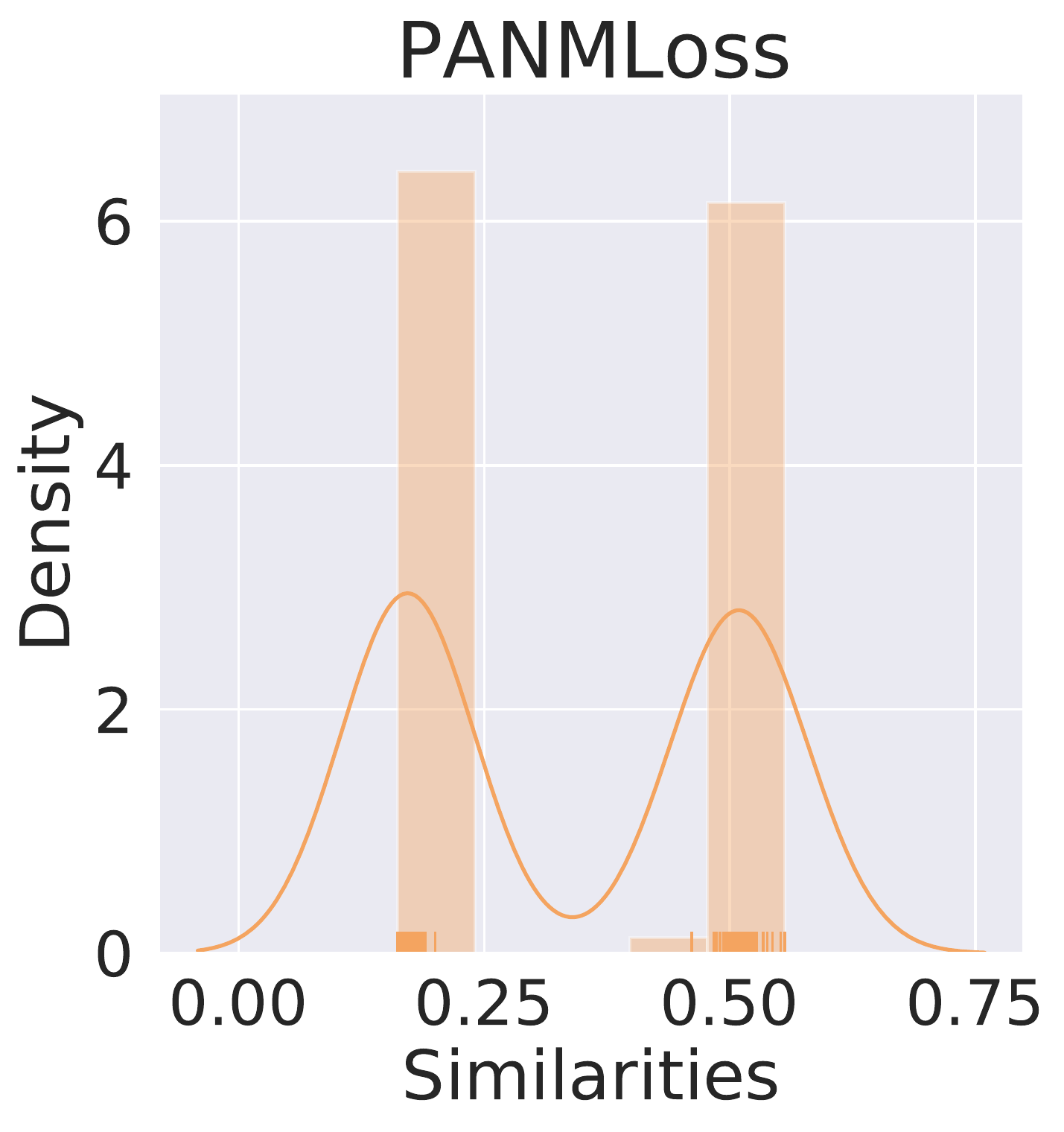}\hspace{6mm}
\includegraphics[width=0.385\columnwidth]{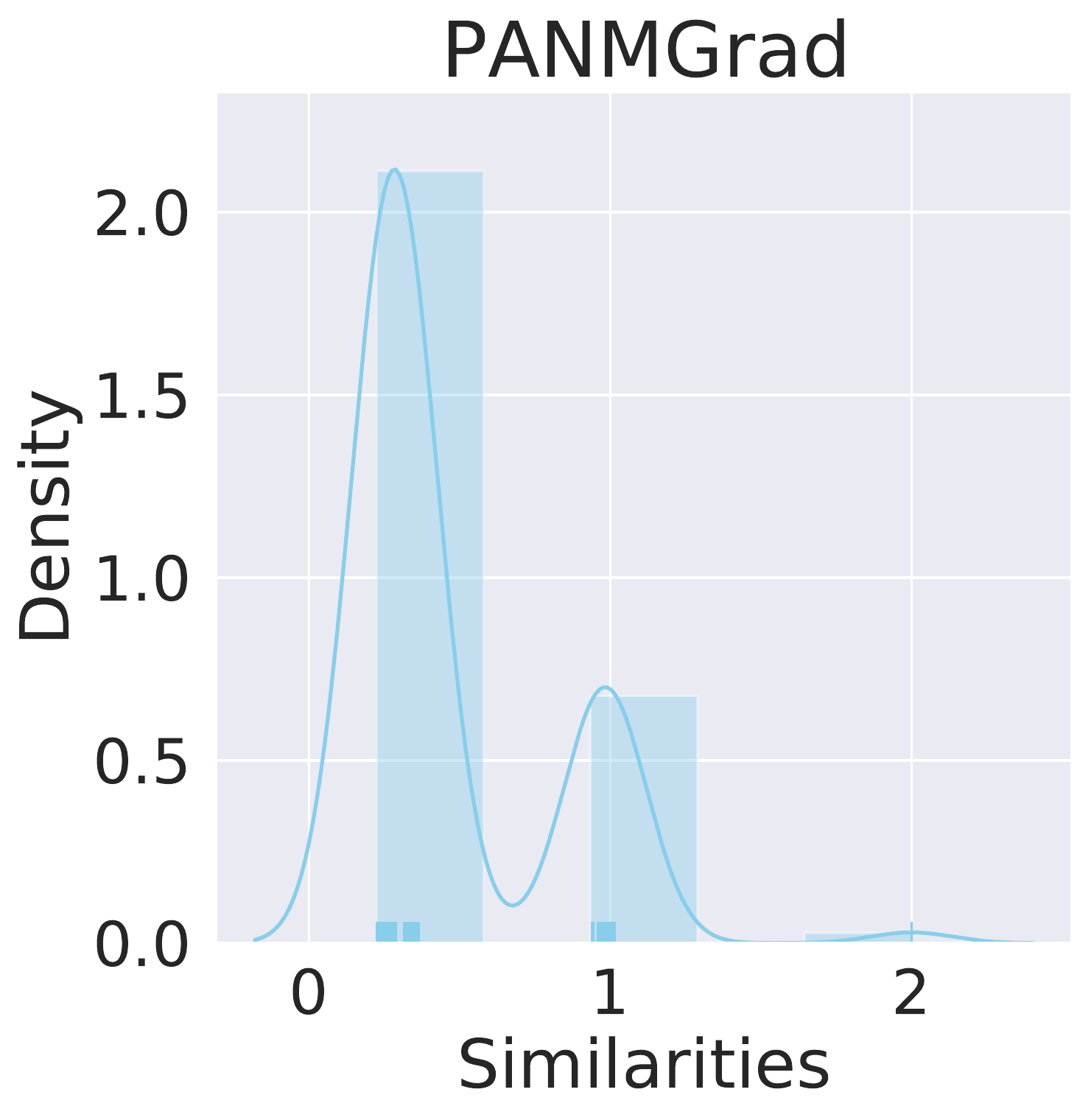}
\caption{Distributions of similarities. Similarities between client 1 and other clients are shown. CIFAR10, $n=100$, clusters are formed by rotations \{0°, 180°\}.}
\label{lossdistri}
\end{figure}

Assumption \ref{assump2} is quite natural in clustered FL. In Figure \ref{lossdistri}, intuitively, the distributions of similarities satisfy our assumption that there are two distinct Gaussians. 

Under Assumption \ref{assump2}, we can implement neighbor augmentation by solving the GMM problem and a typical solution to GMM problem is the EM algorithm. But conventional EM algorithm is not suitable to solve this problem under the following considerations: 
(1) Conventional EM method requires calculating probabilities of all data points in one EM step, but for P2P FL, clients only can communicate with several neighbors in one round. (2) The focus of conventional EM solving GMM problems is to accurately estimate the parameters of Gaussians while our focus is to accurately discriminate cluster identities. (3) Additionally, EM algorithms are sensitive to initialization, and poor initialization may lead to bad convergence.

To tackle above mentioned matters, we devise our Neighbor Augmentation Based on EM-GMM (\texttt{\textbf{NAEM}}) algorithm. In each round, client $i$ randomly samples neighbor candidate list ${\rm C}_{i}^{t} (|{\rm C}_{i}^{t}|=l)$ from non-neighbor clients and also samples a selected neighbor list ${\rm S}_{i}^{t} (|{\rm S}_{i}^{t}|=l$\ if\ $|{\rm B}_{i}^{t}|>l$,\ else\ ${\rm S}_{i}^{t}={\rm B}_{i}^{t})$ from the neighbor bag ${\rm B}_{i}^{t}$. Then, client $i$ communicates with these clients and compute similarities $y_{j}={\rm s}_{i,j},\ j \in {\rm M}_{i}^{t}={\rm C}_{i}^{t} \cup {\rm S}_{i}^{t}$. According to Assumption \ref{assump2}, there are two Gaussian distributions in these similarities, the one with higher mean center refers to the same-cluster clients ($\mathcal N(\mu_{0},\sigma_{0}^{2})$), another one refers to the different-cluster ($\mathcal N(\mu_{1},\sigma_{1}^{2})$). Assuming the observed similarity $y_{j},\ j \in {\rm M}_{i}^{t}$ is generated by the Gaussian Mixture Model:
\begin{equation} \label{equa12}
{\rm Pr}(y|\Theta) = \sum\limits_{r=0}^{1}\beta_{r}\phi(y|\Theta_{r}).
\end{equation}
Here, $\beta_{r}$ refers to the overall probability that $y$ is generated by distribution $r$, and $\Theta=(\beta_{0},\beta_{1};\Theta_{0},\Theta_{1})$. Our target is using EM algorithm to estimate the distribution identities of $y_{j}$, given by 
$$ \gamma_{j,r}=\left\{
\begin{aligned}
1,\ & \text{if}\ j\ \text{belongs\ to\ distribution}\ \mathcal N_{r}; \\
0,\ & \text{otherwise}.\\
\end{aligned}
\right.
$$
where $j \in {\rm M}_{i}^{t},\ r \in \{0, 1\}$. Knowing that EM algorithm is sensitive to initialization, with the prior knowledge that most of the clients in ${\rm S}_{i}^{t}$ are true neighbors, so we can initialize a better parameter as 
$$ \gamma_{j,r}^{(1)}=\left\{
\begin{aligned}
1,\ & j\in {\rm S}_{i}^{t}\ \text{and}\ r=1,\ \text{or}\ j\in {\rm C}_{i}^{t}\ \text{and}\ r=0; \\
0,\ & \text{otherwise}.\\
\end{aligned}
\right.
$$
While the latent variable is the distribution parameters: $\Theta_{0}=(\mu_{0},\sigma_{0}), \Theta_{1}=(\mu_{1},\sigma_{1})$, so the complete data is
\begin{equation} 
(y_{j}, \Theta_{0}, \Theta_{1}),\ j\in {\rm M}_{i}^{t}.
\nonumber
\end{equation}
Then we formulate the expectation function $Q$, based on the log likelihood function of complete data,
\begin{equation} \label{equa13}
\begin{split}
    &Q(\gamma,\gamma^{(a)}) = \mathbb{E}[\log{\rm Pr}(y,\Theta|\gamma)|y,\gamma^{(a)}]\\
    &=\sum\limits_{r=0}^{1}\bigg\{
    n_{r}\log\mathbb{E}\beta_{r}+\sum\limits_{j\in {\rm M}_{i}^{t}} \gamma_{j,r}[\log(\frac{1}{\sqrt{2\pi}})-\log\mathbb{E}\sigma_{r}\\
    &-\frac{1}{2\mathbb{E}\sigma_{r}^{2}}(y_{j}-\mathbb{E}\mu_{r})^{2}]\bigg\},
\end{split}
\end{equation}
where $n_{r}=\sum\limits_{j \in {\rm M}_{i}^{t}}\gamma_{j,r}$, $a$ denotes the iteration step. 

\textbf{E-step:} Now we need to estimate $\mathbb{E}(\mu_{r},\sigma_{r},\beta_{r})$, notated as $\hat{\mu}_{r},\ \hat{\sigma}_{r},\ \hat{\beta}_{r}$.
\begin{equation}
\begin{split}
\hat{\mu}_{r} = \frac{\sum\limits_{j \in {\rm M}_{i}^{t}} \gamma_{j,r}y_{j}}{n_{r}}, \hat{\beta}_{r} = \frac{n_{r}}{|{\rm M}_{i}^{t}|},\hat{\sigma}_{r}^{2} = \frac{\sum\limits_{j \in {\rm M}_{i}^{t}} \gamma_{j,r}(y_{j}-\hat{\mu}_{r})^{2}}{n_{r}},
\end{split}
\nonumber
\end{equation}
where $ r\in\{0,1\}$.

\textbf{M-step:} Iterative M-step is to find the maximum of the function $Q(\gamma,\gamma^{(a)})$ with respect to $\gamma^{(a)}$, as to set $\gamma^{(a+1)}$ in the next iterative epoch
\begin{equation}
\gamma^{(a+1)} = \mathop{\arg\max}_{\gamma} Q(\gamma,\gamma^{(a)}).
\end{equation}
We use the following function to maximize expectation, since $y_{j}$ more likely belongs to $\mathcal N_{0}$ if $\beta_{0}\phi(y_{j}|\Theta_{0})>\beta_{1}\phi(y_{j}|\Theta_{1})$, vice versa.
\begin{equation}
\begin{split}
\gamma_{j,r}^{(a+1)} = &
\mathbbm{1}\Bigg\{ r = \mathop{\arg\max}_{r}
\frac{\hat{\beta}_{r}\phi(y_{j}|\hat{\Theta}_{r})}
{\sum\limits_{c=0}^{1}\hat{\beta}_{c}\phi(y_{j}|\hat{\Theta}_{c})}
\Bigg\},\\
j \in& {\rm M}_{i}^{t}, r \in \{0,1\}.
\end{split}
\end{equation}

Repeat the E-step and M-step until $\gamma^{(a+1)}=\gamma^{(a)}$. Then we obtain the estimated true neighbors in this round notated as ${\rm H}_{i}^{t}$, where $\gamma_{j,0}=1,\ j \in {\rm H}_{i}^{t},\ {\rm H}_{i}^{t}\subseteqq {\rm M}_{i}^{t}$, then we update the neighbor bag, 
\begin{equation}
{\rm B}_{i}^{t+1} = ({\rm B}_{i}^{t}-{\rm S}_{i}^{t})\cup {\rm H}_{i}^{t}.
\end{equation}
By the \texttt{NAEM} algorithm, clients can continually update their neighbor bags, adding new same-cluster peers and removing outliers in the neighbor bag. For model aggregation, clients can conduct gossip communication with the peers in the neighbor bag.

\subsection{\texttt{\textbf{PANM}}: Personalized Adaptive Neighbor Matching} \label{subsect:panm}
\ 
\newline
Now we present \texttt{PANM} by combining the algorithms mentioned above. In the first stage, client $i\ (i \in [n])$ communicates randomly in the network while conducting \texttt{NSMC} for $T_{1}$ rounds. The neighbor list ${\rm N}_{i}^{t}$ in the last round of the first stage is set as the initial neighbor bag in the second stage, ${\rm B}_{i}^{T_1+1} = {\rm N}_{i}^{T_1}$. In the second stage, client $i$ operates gossip communication with peers ${\rm N}_{i}^{t}$ (${\rm N}_{i}^{t}\subseteqq {\rm B}_{i}^{t}, |{\rm N}_{i}^{t}|=k$) sampled from the neighbor bag for aggregation and performs \texttt{NAEM} every $\tau$ rounds to update the neighbor bag ${\rm B}_{i}^{t}$. The process of \texttt{PANM} is shown in Algorithm \ref{PANM}. 

\begin{table*}[!ht]
\normalsize
\caption{\small Complexity analysis regarding computation and communication. Please refer to Table \ref{tab:notation} for the meaning of notations. }
\centering
\begin{tabular}{c|ccc}
\toprule
Methods &Similarity Computation&Communication cost&Maximal required bandwidth\\
\midrule
\texttt{FedAvg }& $\backslash$ &$\mathcal{O}(n(T_{1}+T_{2})$ &$\mathcal{O}(n)$  \\
\texttt{IFCA} &$\mathcal{O}(dcn(T_{1}+T_{2}))$ &$\mathcal{O}(n(c+1)(T_{1}+T_{2}))$ &$\mathcal{O}(cn)$  \\
\texttt{PENS} &$\mathcal{O}(dlnT_{1}+dknT_{2})$ &$\mathcal{O}(nlT_{1}+nkT_{2})$ &$\mathcal{O}(l)$  \\
\midrule
\texttt{PANMLoss} &$\mathcal{O}(d(k+l)nT_{1}+dn(l\tau+kT_{2}))$ &$\mathcal{O}(n(l+k)T_{1}+n(l\tau+kT_{2}))$ &$\mathcal{O}(l+k)$  \\
\texttt{PANMGrad} &$\mathcal{O}((k+l)nT_{1}+n(l\tau+kT_{2}))$ &$\mathcal{O}(n(l+k)T_{1}+n(l\tau+kT_{2}))$ &$\mathcal{O}(l+k)$  \\
\bottomrule
\end{tabular}
\label{tabl:complexity}%添加标题 设置标签
\end{table*}
\noindent\textbf{Complexity Analysis of \texttt{PANM}. }We conduct complexity analysis regarding similarity computation\footnote{We note that the EM steps in \texttt{NAEM} require much little computation because: (1) the computation of similarity measurement includes the operation of high-dimensional vectors, while the EM steps only operate a small set of scalars (the similarity values); (2) the number of similarity values is small; (3) due to our better initialization, it is faster to converge. Thus, we omit the computation of EM steps in the analysis.}, communication cost, and maximal required bandwidth as shown in Table \ref{tabl:complexity}. In Table \ref{tabl:complexity}, \texttt{FedAvg} and \texttt{IFCA} are centralized FL algorithms, and \texttt{IFCA} is the state-of-the-art clustered FL approach. \texttt{PENS} and our \texttt{PANMLoss} and \texttt{PANMGrad} are P2P FL methods. For the similarity computation, the complexity of the centralized \texttt{IFCA} relies on the assumed number of clusters ($c$), while the P2P methods rely on the number of communicated peers ($l, k$). We notice the computation is much more efficient in \texttt{PANMGrad} since it does not rely on local data inference. For communication cost, \texttt{IFCA} has more overhead than \texttt{FedAvg}, and in a sparse P2P network where $l \approx c$, the P2P methods have a similar overhead to \texttt{IFCA}. As for maximal required bandwidth, P2P approaches have a dominant advantage over the centralized since $n \gg (l+k)$ is usually held in practice. We notice that \texttt{PANM} has more overhead compared with \texttt{PENS} in the first stage, but \texttt{PANM} realizes a more pure neighbor selection, that we can set a relatively smaller $T_{1}$ for \texttt{PANM} to reduce the overhead.

\noindent\textbf{Applying \texttt{PANM} under Server-client Protocol. }In applications, if only the server-client protocol is available, \texttt{PANM} can still be applied by simply taking the server as a relay for transmitting models. Concretely, clients send local models and requests about which peers it wants to communicate with to the server, and then the server sends the requested models to corresponding clients. The similarity measurement and aggregation process are implemented on the client side. Clients update their neighbor lists and generate the request in each round. The scheme that the server sends other clients' models to a client for updating the client similarity matrix is adopted in personalized FL \cite{zhang2020personalized}, showing this scheme is realistic in practice. Nevertheless, our method is essentially a P2P algorithm where clients have the autonomy to choose the neighbors, while in \cite{zhang2020personalized}, the server updates the client similarity matrix and has the autonomy.

% PANM
\begin{algorithm}[th] 
  \caption{\small \texttt{PANM}: Personalized Adaptive neighbor Matching}
  \textbf{Input}: {\small $n,k,l,T_{1},T_{2},\eta, E, \tau,\alpha,\mathbf{w}_{0},\mathbf{W}^{0}=\{ \mathbf{w}_{0}^{1}=\mathbf{w}_{0},i\in[n] \}$;}\\
  \textbf{Output}: $\mathbf{W}^{T_{1}+T_{2}}$, {\rm B};
  
  \begin{algorithmic}[1]
  \STATE Initiate neighbor list: ${\rm N}_{i}^0$;
    \FOR{each round $t=1,\dots, T_{1}+T_{2}$}
        \FOR{each client $i, i\in[n]$ \textbf{in parallel}}
        \STATE Compute $E$ epochs of local training:
        \STATE $\mathbf{w}_{i}^{t-\frac{1}{2}}\gets\mathbf{w}_{i}^{t-1}-\eta\nabla F_{i}\left(\mathbf{w}_{i}^{t-1}\right)$;
        \IF {$t\in [T_{1}]$}
            \STATE ${\rm N}_{i}^{t}\gets$\texttt{NSMC}(${\rm N}_{i}^{t-1}$),
            \STATE $\mathbf{w}_{i}^{t}\gets$ Aggregation(${\rm N}_{i}^{t}$,$\mathbf{w}_{i}^{t-\frac{1}{2}}$);
        \ELSE 
        \STATE ${\rm B}_{i}^{T_1+1}={\rm N}_{i}^{T_1}$;
        \IF {$t \% \tau = 0$}
        \STATE ${\rm B}_{i}^{t}\gets$ \texttt{NAEM}(${\rm B}_{i}^{t-1}$),
        \STATE ${\rm N}_{i}^{t}\gets$ RandomSample(${\rm B}_{i}^{t}$),
        \STATE $\mathbf{w}_{i}^{t}\gets$ Aggregation(${\rm N}_{i}^{t}$,$\mathbf{w}_{i}^{t-\frac{1}{2}}$);
        \ELSE 
        \STATE ${\rm B}_{i}^{t}\gets{\rm B}_{i}^{t-1}$,
        \STATE ${\rm N}_{i}^{t}\gets$ RandomSample(${\rm B}_{i}^{t}$),
        \STATE $\mathbf{w}_{i}^{t}\gets$ Aggregation(${\rm N}_{i}^{t}$,$\mathbf{w}_{i}^{t-\frac{1}{2}}$);
        \ENDIF
        \ENDIF
        \ENDFOR
    \ENDFOR
  \end{algorithmic}
\label{PANM}
\end{algorithm}

\subsection{Theoretical Analysis} \label{subsect:theory}
In this section, we provide theoretical analysis. We first give the probability model about the expected probability that all neighbors are true in Theorem \ref{thm1}. It shows that the Monte Carlo method will enable \texttt{PANM} to have a increasingly pure neighbors in the first stage, while \texttt{PENS} has constant expectation on the purity. Then, we provide a unified framework about the one-round error bound to the clustered optimum in Theorem \ref{thm2}. In conjunction with Theorems \ref{thm1} and \ref{thm2}, we deduce the error bound of \texttt{PANM} in Theorem \ref{thm3}. 

\noindent\textbf{Probability Model about the Purity of Neighbors.} We first deduce the probability model by providing the assumption about the effectiveness of similarity measurement.
\begin{assumption} \label{assump1}
\textbf{(Effectiveness of Similarity Measurement)} The metrics in Equation \ref{equ6} and \ref{equ5} are effective enough, so that for client $i$ ($\forall i \in [n]$), we have:
\begin{equation} \label{equ9}
{\rm s}_{i,p} > {\rm s}_{i,q}\quad \forall p \in {\rm N}_{i}^*,\ q \in \overline{{\rm N}_{i}^*}.
\end{equation}
\end{assumption}
Given Assumption \ref{assump1}, we can provide the expected probability that all neighbors are true in round $t$. Recall that there are $n$ clients in the system (including client $i$) and $a$ clients in the same cluster as client $i$ (including client $i$), so we can infer the following theorem.
\newtheorem{thm}{Theorem}
\begin{thm} \label{thm1}
\textbf{(Expected Probability of True Neighbors)} Under Assumption \ref{assump1}, in the round $t$ when conducting \texttt{NSMC} as Equation \ref{equ8} where ${\rm L} = {\rm N}_{i}^{t-1}$ when $t > 1$; for client $i$, the expected probability that all neighbors are true is ${\rm P}^{t}(k)$, we have
\begin{equation} \label{equ10}
\begin{split}
{\rm P}^{t}(k) &= {\rm G}(k)\ast {\rm P}^{t-1}(k) + {\rm R}(k)\\
&\dots\\
{\rm P}^{2}(k) &= {\rm G}(k)\ast {\rm P}^{1}(k) + {\rm R}(k)\\
{\rm P}^{1}(k) &= {\rm R}(k).
\end{split}
\end{equation}
where {\rm R}(x) and {\rm G}(x) are two functions and the '$\ast$' refers to the discrete convolution computation, defined as 
\begin{small}
\begin{equation}
\begin{split}
&{\rm R}(x) = \small{\frac{l!(n-l-1)!}{(n-1)!}\sum\limits_{s=0}^{l-x} \frac{(a-1)!(n-a)!}{s!(l-s)!(n-a-s)!(a-l+s-1)!}}\\
&{\rm G}(x) = \frac{l!(a-1)!(n-a)!(n-l-1)!}{x!(n-1)!(l-x)!(a-x-1)!(n-a-l+x)!}\\
&{\rm G}(x)\ast {\rm P}(x) = \sum\limits^{x-1}_{m=0} {\rm G}(m){\rm P}(x-m).
\nonumber
\end{split}
\end{equation}
\end{small}
If conducting \texttt{PENS} as Equation \ref{equ8} where ${\rm L} = \emptyset$, the expected probability is 
\begin{equation}
    {\rm P}^{t}(k) \equiv {\rm R}(k).
\end{equation}
\end{thm}
Based on Theorem \ref{thm1}, we provide the following corollary.
\newtheorem{corollary}{Corollary}
\begin{corollary} \label{corollary1}
\textbf{(The Monotonicity of Probability Functions)} Given Theorem \ref{thm1}, we denote ${\rm Q}(t) = {\rm P}^{t}(k), t\in [T]$ as the probability function of round $t$. The function ${\rm Q}(t)$ of \texttt{NSMC} is monotone increasing, so we have \\
\begin{equation} 
    {\rm Q}(t)>{\rm Q}(t-1)>\dots>{\rm Q}(2)>{\rm Q}(1).
\end{equation}
The function ${\rm Q}(t)$ of \texttt{PENS} is constant which satisfies
\begin{equation}
    {\rm Q}(t)={\rm Q}(t-1)=\dots={\rm Q}(1)={\rm R}(k).
\end{equation}
\end{corollary}

The proofs of Theorem \ref{thm1} and Corollary \ref{corollary1} are shown in the appendix. The Corollary \ref{corollary1} shows that: by \texttt{NSMC}, the probability increases over round, while \texttt{PENS} keeps it unchanged at a low value. Intuitively, we calculate the theoretical probability under different settings in Table \ref{tab:confidence}. It is obvious that the probability of \texttt{NSMC} increases fast, it will reach 100\% in round 5; whereas \texttt{PENS} will have constantly low probability. 
\begin{table}[h]
\caption{Theoretical precision of true neighbors under different settings. The results are shown in percentage (\%).}
\centering
\begin{tabular}{c|cc|cc|cc}
\toprule
&\multicolumn{2}{|c}{t=3}&\multicolumn{2}{|c}{t=5}&\multicolumn{2}{|c}{t=7}\\ 
% \cline{2-7}
\midrule
n, a, l, k & \texttt{PENS} &\texttt{NSMC}& \texttt{PENS} &\texttt{NSMC}& \texttt{PENS} &\texttt{NSMC}\\
% \cline{2-13}
\midrule
200,50,10,5&	7.29&	90.75&	7.29&	99.82&	7.29&	100.00 \\
200,50,20,10&	0.98&	96.24&	0.98&	100.00&	0.98&	100.00 \\
200,50,20,6&	38.00&	99.94&	38.00&	100.00&	38.00&	100.00 \\
100,50,10,5&	62.97&	100.00&	62.97&	100.00&	62.97&	100.00 \\
\bottomrule
\end{tabular}
\label{tab:confidence}
\end{table}
However, we note that Assumption\ref{assump1} is strong, and we provide this assumption just for theoretical analysis. There are gaps between theory and practice, especially in the first couple of rounds, when the similarity measurement is not effective enough. If the effectiveness of similarity is not strongly hold as Assumption\ref{assump1}, the probability in Theorem \ref{thm1} will have a discount, but the monotone increasing property of \texttt{NSMC} in Corollary \ref{corollary1} still holds.

\noindent\textbf{Error Bound to the Clustered Optimum.} Inspired by \cite{DBLP:conf/nips/GhoshCYR20}, we propose 
a general theoretical framework to analyse the convergence and error bound under clustered heterogeneity in P2P FL. Then we incorporate Theorem \ref{thm1} to give the error bound of \texttt{PANM}. To start with, we first give the following definitions.

\newtheorem{definition}{Definition}
\begin{definition} \label{definit1}
\textbf{(Optimality within a Cluster)} Knowing that there are $r$ clusters, for the data distribution of cluster $j$ ($j \in [r]$), we define the population loss of cluster $j$ and the optimal model of cluster $j$ as
\begin{equation}
\begin{split}
    F^{j}(\mathbf{w})&\coloneqq\mathbb{E}_{\xi \sim\mathcal{D}^{j}}\left[f(\mathbf{w}_{i},\xi_{i})\right],\\
    \mathbf{w}_{j}^{*} &= \arg\min_{\mathbf{w}} F^{j}(\mathbf{w}).
    \nonumber
\end{split}
\end{equation}

\end{definition}
\begin{definition} \label{definit2}
\textbf{(Cluster Heterogeneity)} We define the cluster heterogeneity as the maximal distance between the optimal models of each cluster.
\begin{equation}
    \Delta \coloneqq \max_{i \neq j; i,j \in [r]}\Vert \mathbf{w}_{i}^{*} - \mathbf{w}_{j}^{*}\Vert.
    \nonumber
\end{equation}
\end{definition}

\begin{definition} \label{definit3}
\textbf{(The Error Rate of Neighbor Estimation)} For a given client $i$, the error rate in the aggregation neighbor list is the proportion of false neighbors to all neighbors, 
\begin{equation}
    \epsilon = \frac{| {\rm N}_{i} \cap \overline{{\rm N}_{i}^{*}} |}{| {\rm N}_{i} |} = \frac{| {\rm N}_{i} \cap \overline{{\rm N}_{i}^{*}} |}{k}.
    \nonumber
\end{equation}
Thus, the number of true neighbors is $| {\rm N}_{i} \cap {\rm N}_{i}^{*} | = k(1 - \epsilon)$, and the number of false neighbors is $| {\rm N}_{i} \cap \overline{{\rm N}_{i}^{*}} | = k\epsilon$.
\end{definition}

\noindent Then, we give the following assumptions for theoretical analysis.
\begin{assumption} \label{assmp_for_thm1}
\textbf{($\mu$-strongly Convexity)} The loss functions of each client $F_{i}(\mathbf{w}), \forall i \in [n]$ and each cluster $F^{j}(\mathbf{w}), \forall j \in [r]$ are all $\mu$-strongly convex that satisfy: $\forall \mathbf{w}, \mathbf{w}^{\prime}$,
\begin{equation}
    F(\mathbf{w}^{\prime}) \geq F(\mathbf{w}) + \langle \nabla F(\mathbf{w}), \mathbf{w}^{\prime}-\mathbf{w}\rangle + \frac{\mu}{2}\Vert \mathbf{w}^{\prime} - \mathbf{w}\Vert^{2}.
    \nonumber
\end{equation}
\end{assumption}

\begin{assumption} \label{assmp_for_thm2}
\textbf{($L$-smoothness)} The loss functions of each client $F_{i}(\mathbf{w}), \forall i \in [n]$ and each cluster $F^{j}(\mathbf{w}), \forall j \in [r]$ are all $L$-smooth that satisfy: $\forall \mathbf{w}, \mathbf{w}^{\prime}$,
\begin{equation}
    F(\mathbf{w}^{\prime}) \leq F(\mathbf{w}) + \langle \nabla F(\mathbf{w}), \mathbf{w}^{\prime}-\mathbf{w}\rangle + \frac{L}{2}\Vert \mathbf{w}^{\prime} - \mathbf{w}\Vert^{2}.
    \nonumber
\end{equation}
\end{assumption}

\begin{assumption} \label{assmp_for_thm3}
\textbf{(Bounded Gradient Variance)} We bound the variance of gradients within a cluster. For every $\mathbf{w}$ and every $j \in [r]$, the variance of $\nabla f(\mathbf{w},\xi)$ is upper bounded by $v^2$, when $\xi$ is sampled from $\mathcal{D}^{j}$.
\begin{equation}
    \mathbb{E}_{\xi \sim\mathcal{D}^{j}}[\Vert \nabla f(\mathbf{w},\xi) - \nabla F^{j}(\mathbf{w}) \Vert^{2}] \leq v^2.
    \nonumber
\end{equation}
\end{assumption}

\noindent Given the above definitions and assumptions, we present the theorem about the error bound within one communication round in arbitrary P2P algorithms. 
\begin{thm} \label{thm2}
\textbf{(Error Bound within One Communication Round)} For a client $i, i \in [n]$, which belongs to cluster $j, j \in [r]$, in a certain communication round, the error gap between its model to the clustered optimum is $\Vert \mathbf{w}_{i} - \mathbf{w}_{j}^{*} \Vert$. Let $\mathbf{w}_{i}^{+}$ be the next-round model after communicating with neighbors. Then, the next-round error gap is 
\begin{equation} \label{thm1:equ}
\begin{split}
    \Vert \mathbf{w}_{i}^{+} - \mathbf{w}_{j}^{*} \Vert &\leq (1-\frac{\eta \mu L(1-\epsilon)}{\mu + L} + \eta L \epsilon)\Vert \mathbf{w}_{i} - \mathbf{w}_{j}^{*} \Vert \\
    &+ \eta L \Delta \epsilon + \frac{v}{\sqrt{dk}}\frac{1}{\sqrt{1-\epsilon}} + \eta v \sqrt{\frac{r}{kd}}\sqrt{\epsilon}.
\end{split}
\end{equation}
\end{thm}

\newtheorem{remark}{Remark}
\begin{remark} \label{remark}
\textbf{(Error Bound under Different $\epsilon$)} Note that Theorem \ref{thm2} is a unified bound for any algorithm and the differences between algorithms lie in the error rate of neighbor estimation $\epsilon$.

If the algorithm is effective enough that have $\epsilon \rightarrow 0$, we have
\begin{equation} 
        \Vert \mathbf{w}_{i}^{+} - \mathbf{w}_{j}^{*} \Vert \leq (1 - \frac{\eta \mu L}{\mu + L}) \Vert \mathbf{w}_{i} - \mathbf{w}_{j}^{*} \Vert +\frac{v}{\sqrt{dk}}.
\end{equation}
In this case, it is clear that the $\Vert \mathbf{w}_{i} - \mathbf{w}_{j}^{*} \Vert$ term is decreasing, and if $d, k$ are large while $v$ is small, the model will converge to the clustered optimum.

If the algorithm is dump that $\epsilon \rightarrow \frac{k-1}{k}$, we have 
\begin{equation}
        \Vert \mathbf{w}_{i}^{+} - \mathbf{w}_{j}^{*} \Vert \leq (1 + \eta L) \Vert \mathbf{w}_{i} - \mathbf{w}_{j}^{*} \Vert + \eta L \Delta + \eta v \sqrt{\frac{r}{kd}} +\frac{v}{\sqrt{d}}.
\end{equation}
In this case, $(1 + \eta L)$ shows that it is not converging to the clustered optimum. If the clustered heterogeneity is more dominant with larger $\Delta$, the model will be further away from the optimum.

From Equation \ref{thm1:equ}, we can infer that if $1-\frac{\eta \mu L(1-\epsilon)}{\mu + L} + \eta L \epsilon > 1 \Leftrightarrow \epsilon > \frac{\mu}{2\mu+L}$, the convergence to the clustered optimum is not satisfied. More intuitively, if we assume $\mu = L$, the condition becomes $\epsilon > \frac{1}{3}$, which means if the proportion of false neighbors is larger than $\frac{1}{3}$ in each round, it is impossible to converge to the clustered optimal. Therefore, the neighbor estimation and selection is quite essential for clients to converge to the clustered optimum.
\end{remark}

\begin{thm} \label{thm3}
\textbf{(Error Bound of \texttt{PANM} in the First Stage)} We assume Assumptions \ref{assump2}-\ref{assmp_for_thm3} hold and $\mu = L$, and set the learning rate $\eta = \frac{1}{L}$. We analyse the error bound of client $i, i\in [n]$, which belongs to cluster $j, j\in [r]$, when applying \texttt{PANM} in the P2P FL system. The initial error gap is defined as $\delta_{0} = \Vert \mathbf{w}_i^0 - \mathbf{w}_j^* \Vert$. After $T$ rounds, the error bound is 
\begin{equation}
\begin{split}
    \Vert \mathbf{w}_{i}^{T} - \mathbf{w}_{j}^{*} \Vert &\leq \frac{1}{2^{T-1}}	{\Bigg[}\frac{1+3\epsilon_{0}}{2}\delta_{0} + \epsilon_{0}\Delta +\frac{v}{\sqrt{dk(1-\epsilon_{0})}} \\
    &+\frac{v}{L}\sqrt{\frac{r\epsilon_{0}}{kd}}\Bigg] + \sum_{t=0}^{T-2}\frac{1}{2^t}\frac{v}{\sqrt{dk}},
\end{split}
\end{equation}
where $\epsilon_{0} = {\rm R}(k) = \small{\frac{l!(n-l-1)!}{(n-1)!}\sum\limits_{s=0}^{l-k} \frac{(a-1)!(n-a)!}{s!(l-s)!(n-a-s)!(a-l+s-1)!}}$.
\end{thm}
The proofs of Theorems \ref{thm2} and \ref{thm3} are shown in the appendix. From Theorems \ref{thm2} and \ref{thm3}, we prove that \texttt{PANM} can converge to the clustered optimum and it has a linear convergence rate. It is worthy mentioning that the effect of clustered heterogeneity $\Delta$. Intuitively, more dominant heterogeneity will result in more effective similarity measurement, in other words, Assumption \ref{assump1} is more likely to be held and the error rate of neighbor estimation $\epsilon$ is lower. In Remark \ref{remark}, lower error rate will result in faster convergence to the optimum. However, on the other hand, larger $\Delta$ will have a more dominant error term in Theorem \ref{thm3}. 

If we can formulate $\epsilon$ into a function of $\Delta$, the effect of heterogeneity is more tractable. For instance, if $\epsilon = \mathcal{O}(\frac{1}{\Delta^2})$, the error term $\Delta\epsilon = \mathcal{O}(\frac{1}{\Delta})$, in this case, larger heterogeneity will benefit; and if $\epsilon = \mathcal{O}(\frac{1}{\sqrt{\Delta}})$, the error term $\Delta\epsilon = \mathcal{O}(\sqrt{\Delta})$, thus, larger heterogeneity will lead to larger error gap; and if $\epsilon = 0$ in all rounds, the effect of $\Delta$ will be removed.

\section{Experiments and Results}\label{sect:exp}
In this section, we evaluate our methods and compare them with baselines. P2P FL baselines include \texttt{PENS} \cite{DBLP:journals/corr/abs-2107-08517} (state-of-the-art personalized P2P FL algorithm), \texttt{Random} (gossip with random neighbors), \texttt{Local} (without communication), \texttt{FixTopology} (neighbors are randomly sampled at the beginning and fixed during training). We also include \texttt{Oracle} (with prior knowledge of cluster identities, gossip with true neighbors) for comparison, and it is not a baseline but the ideal gossip algorithm with ground-truth cluster information, which is not realistic in practice. \texttt{Oracle} may indicate the upper bound of accuracy in clustered P2P FL, but the following experiments will show our methods can sometimes surpass it.

Centralized FL baselines include \texttt{IFCA} \cite{DBLP:conf/nips/GhoshCYR20} (state-of-the-art centralized clustered FL) and centralized \texttt{FedAvg} \cite{DBLP:conf/aistats/McMahanMRHA17}. Our methods include \texttt{PANMLoss} (\texttt{PANM} with metric based on loss), \texttt{PANMGrad} (\texttt{PANM} with metric based on weight updates and gradients).

\subsection{Settings of Datasets}
\textbf{Synthetic Clustered Heterogeneity.} We use three public benchmark datasets, MNIST \cite{lecun-mnisthandwrittendigit-2010}, CIFAR10 \cite{krizhevsky2009learning}, and FMNIST (Fashion-MNIST) \cite{xiao2017fashion}. To synthesize clustered heterogeneity, we use rotation transformation and label-swapping, respectively; we note these two settings are commonly used in clustered FL (rotation \cite{DBLP:journals/corr/abs-2107-08517,DBLP:conf/nips/GhoshCYR20}, label-swapping \cite{DBLP:journals/tnn/SattlerMS21}). All results are evaluated on each client's local testset, and we keep the size of the local testset to 100 for all scenarios and present the averaged results among all clients. We then describe the details of clustered rotation transformation and label-swapping. 

Rotation Transformation: There are two settings for rotation transformation. First is rotation with two clusters (\{0°, 180°\}): clients in cluster 0 keep images without any transformation (0° rotation), while clients in cluster 1 rotate every image in the trainset and testset for 180°. The second is rotation with four clusters (\{0°, 90°, 180°, 270°\}): clients in cluster 0 keep images without any transformation (0° rotation), and clients in cluster 1 rotate every image in trainset and testset for 90°, clients in cluster 2 for 180°, and clients in cluster 3 for 270°. The labels of images remain unchanged, and each client's class distributions are balanced.

Swapping Labels: There are two settings for swapping labels. First is forming two clusters by swapping labels: for clients in cluster 0, images labeled as ``0'' are relabeled as ``1'' and images labeled as ``1'' are relabeled as ``0''; while for clients in cluster 1, images labeled as ``6'' are relabeled as ``7'' and images labeled as ``7'' are relabeled as ``6''. The second is forming four clusters by swapping labels: (1) for clients in cluster 0, images labeled as ``0'' and ``1'' are swapped by labels; (2) for clients in cluster 1, images labeled as ``2'' and ``3'' are swapped; (3) for clients in cluster 2, images labeled as ``4'' and ``5'' are swapped; (4) for clients in cluster 3, images labeled as ``6'' and ``7'' are swapped. Note that the class distributions in clients are balanced.

\noindent\textbf{Real-world Clustered Heterogeneity.} We also use Digit-five \cite{zhao2020multi,peng2019federated,luo2021ensemble} to validate the algorithms. Digit-five is a collection of five digital recognition datasets, namely handwritten digits (MNIST) \cite{lecun1998gradient}, digits with colored backgrounds (MNIST-M) \cite{ganin2015unsupervised}, street images of digits (SVHN) \cite{netzer2011reading}, synthetic digits (Synthetic Digits) \cite{ganin2015unsupervised}, and digits from postal services (USPS) \cite{hull1994database}. These datasets are of different domains and modalities, and assigning these datasets to clients can inherently realize clustered heterogeneity. USPS is much smaller than others, so we exclude USPS and select 50000 samples from each of the other four domains as the FL trainset. As a result, there exist four clusters among clients. To enable the training under the same model architecture, we transform the images in different domains to the same size and number of channels. 

\subsection{Details of Implementations}
\textbf{Implementation Environment.} All the experiments are implemented in PyTorch 1.7.1. We have several GPUs for training, including Tesla P40 GPU with 24451MB memory, Quadro RTX 8000 GPU with 48601MB memory, Tesla P100 GPU with 16280MB memory, and Tesla V100 GPU with 16130MB memory. 

\noindent\textbf{Clients' Models.} A three-layer MLP with ReLU activations is adopted as the model for training on MNIST and FMNIST. For CIFAR10, a five-layer convolution neural network model (three convolutional layers followed by two fully connected layers) is used. For Digit-five, we use LeNet-5 \cite{lecun2015lenet}. We do not use data augmentation techniques like flipping and random cropping.

\noindent\textbf{Hyperparameters.} We set the batch size for all experiments to 128, and the number of local epochs in each round is 3. We adopt the learning rate decay strategy used in \texttt{IFCA} \cite{DBLP:conf/nips/GhoshCYR20}. The decay step size is 0.99, which means for each round, the learning rate is set to \textit{the learning rate in the last round} × 0.99. The initial learning rate is set to 0.08 in the first round. We use the SGD optimizer and set SGD momentum to 0.9. We set $T_{1}=100,\ T_{2}=200$ in all experiments. For “\textit{the expected amount of times}” in \texttt{PENS}, we set this hyperparameter to an appropriate value ceil($T_{1}\times(l+k)/n$) for fair comparison. For centralized FL: \texttt{IFCA} and FedAvg, we set full participation of clients in each round. For \texttt{PANMGrad}, we set $\alpha=0.5$ in all experiments.

\noindent\textbf{Result Presentation.} In every setting, we conduct experiments with different random initialization three times and average the results, and the mean results and the standard deviations are shown in the tables and figures. All experimental results are shown in percentage value (\%).

\subsection{Results on Different Datasets}
\begin{figure*}[t]
\centering
\includegraphics[width=2.0\columnwidth]{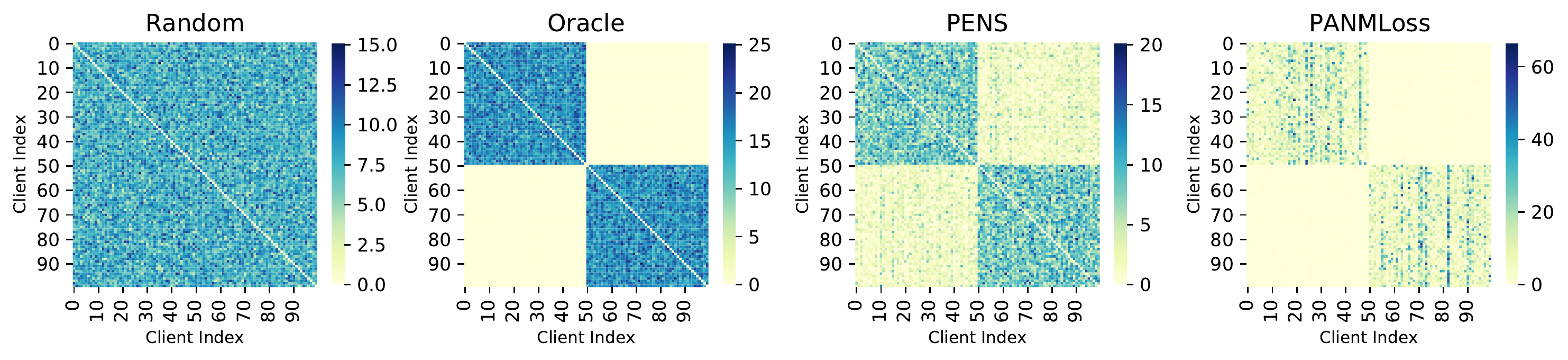} 
\caption{Heatmaps of aggregation records in the first stage. CIFAR-10 with two rotations \{0°:\ clients 0-49,\ 180°:\ clients 50-99\}, trainset size is 400, $l=10,k=5$.}
\label{heatmap}
\end{figure*}
Table \ref{dffdataset} shows the test accuracies of all methods under different datasets. Although \texttt{FixTopology} and \texttt{Random} improve accuracy compared with local training, in contrast to \texttt{Oracle}, the fixed or random topology will impede performance gains. 

\begin{figure}[t]
\subfloat[]{
\includegraphics[width=0.9\columnwidth]{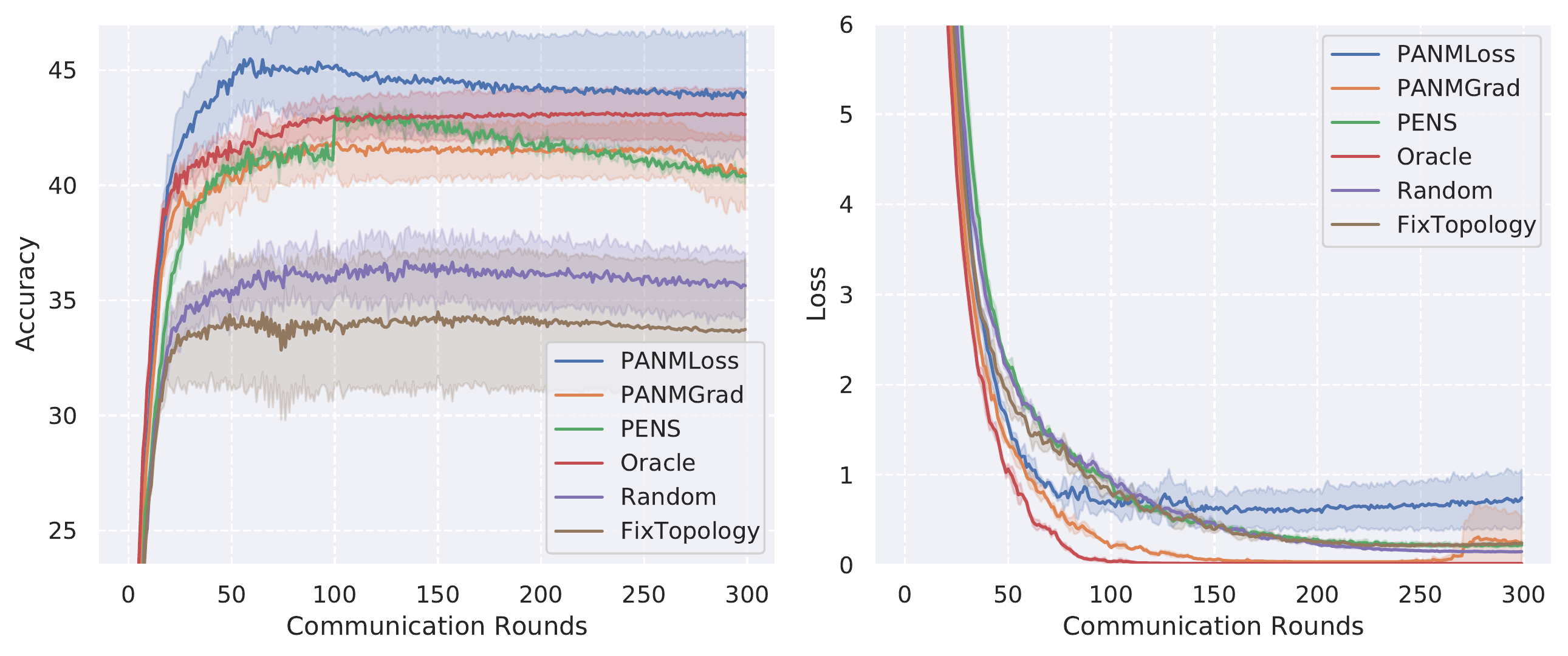} }\\
\subfloat[]{
\includegraphics[width=0.9\columnwidth]{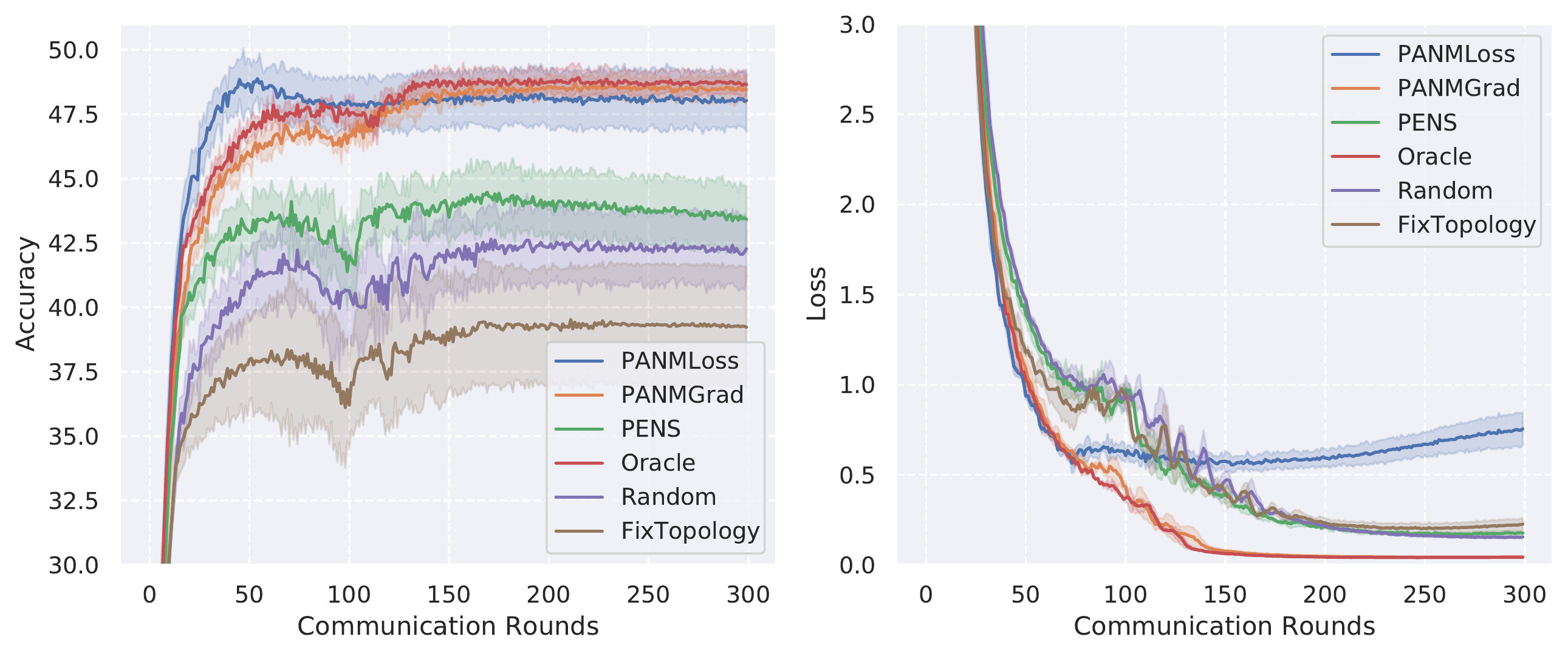} }\\
\subfloat[]{
\includegraphics[width=0.9\columnwidth]{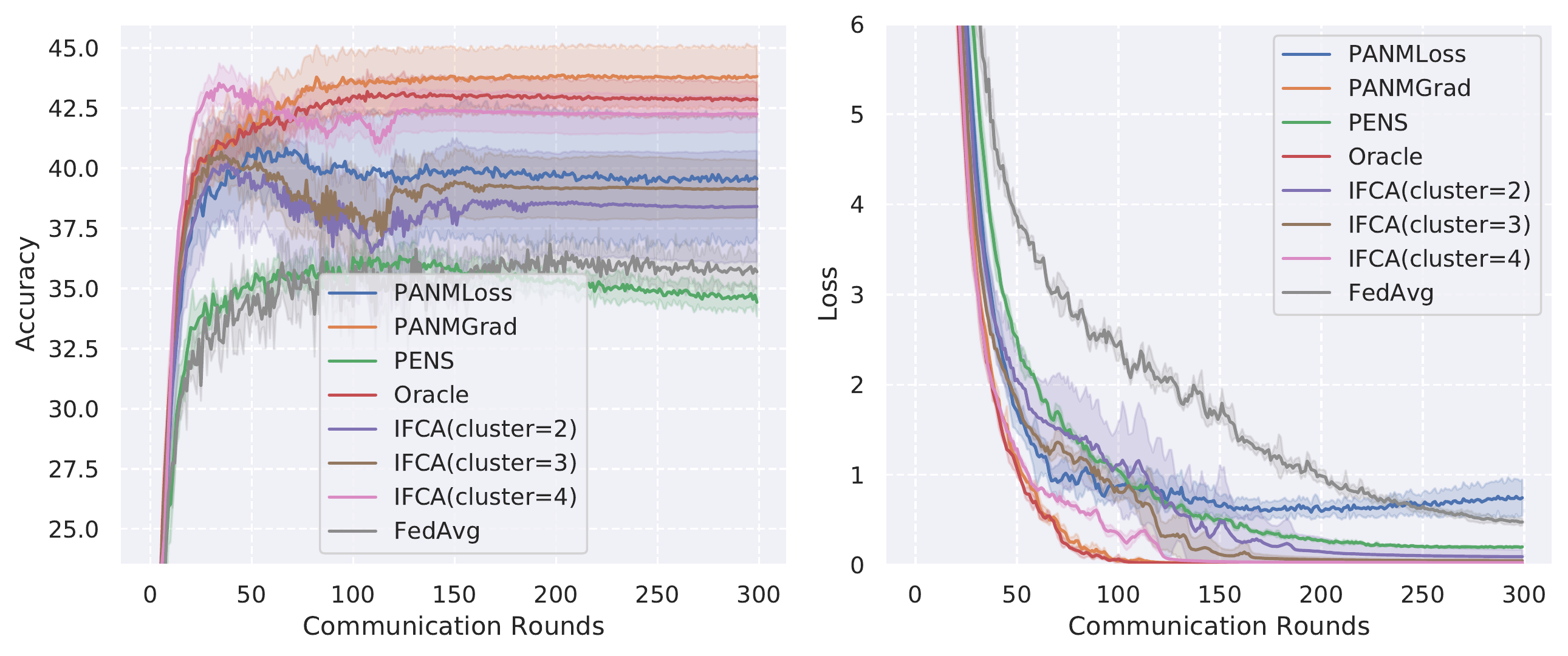} }
\caption{Accuracy and loss curves. CIFAR10, $n=100,l=10,k=5$, trainset size is 200. (a) 4 clusters with swapping labels; (b) 2 clusters with rotations \{0°,180°\}; (c) comparing with centralized FL methods, 4 clusters with rotations \{0°,90°,180°,270°\}.}
\label{acclosscifar}
\end{figure}

Recall \texttt{Oracle} has perfect information about cluster identities which is impossible in real FL scenarios. Notably, our methods surpass \texttt{Oracle} in FMNIST and CIFAR10. We show the accuracy and loss curves of results in CIFAR10 in (a) and (b) of Figure \ref{acclosscifar}. It is demonstrated that \texttt{PANMLoss} achieves high accuracy and fast convergence in the first stage, which means the \texttt{NSMC} algorithm is effective. We explain that the Monte Carlo method enables the clients to collaborate with peers with maximal similarities, so the performance will be better than randomly sampled from same-cluster peers (\texttt{Oracle}). This explanation is also validated in the heatmap of aggregation records between clients in Figure \ref{heatmap}. In \texttt{PANM}, the neighbors are pure compared with \texttt{PENS} and clients are prone to communicate with several peers instead of all true neighbors. 

Additionally, we notice that \texttt{PANM} performs normally in MNIST, and we reckon this is because the clustered heterogeneity $\Delta$ is not dominant in rotated MNIST. One example is that if we rotate the images of ``0'' for 180°, the images represent the same distribution. \texttt{PANM} will not take full advantage when $\Delta$ is small. Firstly, as we have discussed in Section \ref{subsect:theory}, smaller $\Delta$ will make the similarity metrics less effective. More importantly, in this case, the different clustered optimums have close distances, which means that the negative transfer between clusters is weak, so there is less necessity for clustering.

\begin{table}[!t] \small
\caption{Results on different datasets. The top two are in bold. For all datasets: trainset size is 200, two rotations \{0°,180°\}, $l=10,k=5$. $n=100$ for CIFAR10 and FMNIST, $n=200$ for MNIST.}
\centering
\begin{tabular}{c|ccc}
\toprule
Methods&MNIST&FMNIST&CIFAR10\\
\midrule 
Local&82.57 ±0.28 &76.24 ±0.22 &25.27 ±1.21\\
\midrule 
FixTopology&94.71 ±0.09 &85.86 ±0.17 &39.74 ±2.27 \\
Random&95.12 ±0.04 &85.94 ±0.27 &42.96 ±1.42 \\
Oracle&\textbf{95.87 ±0.08} &\textbf{87.01 ±0.26} &\textbf{49.11 ±0.48} \\
PENS&\textbf{96.15 ±0.16} &86.82 ±0.11 &44.78 ±1.12 \\
\midrule
PANMLoss&95.63 ±0.12 &\textbf{87.33 ±0.17} &\textbf{49.19 ±0.79 }\\
PANMGrad&95.65 ±0.09 &86.88 ±0.34 &48.83 ±0.39 \\
\bottomrule
\end{tabular}
\label{dffdataset}
\end{table}

\subsection{Results under Various Clustered Heterogeneity}
Table \ref{dffnoniid} shows results under various heterogeneity: swapping labels and more rotations. For test accuracy, it is evident that our methods are robust in different Non-IID environments. It is notable that in some experiments (CIFAR10 Label-swap(2) and (4)), \texttt{PANM} even outperforms \texttt{Oracle} with a large margin. \texttt{PANMGrad} has similar performances to \texttt{PANMLoss}, but performances vary in different settings. This is due to their different perspectives of client similarity that the loss-based and gradient-based perspectives will take advantage in different scenarios. 
\begin{table}[!t] \small
\caption{Results under various heterogeneity. CIFAR10, $n=100,l=10,k=5$, trainset size is 200. Label-swap(2)/(4): two/four clusters with swapping labels, Rotation(4): four clusters with rotation \{0°,90°,180°,270°\}.}
\centering
\begin{tabular}{c|ccc}
\toprule
\ &\multicolumn{3}{c}{Test Accuracy}\\
\midrule 
Methods&Label-swap(2)&Label-swap(4)&Rotation(4)\\
\midrule 
Local&25.27 ±1.21 &25.27 ±1.21 &25.27 ±1.21\\
\midrule 
FixTopology&36.56 ±1.58 &35.08 ±2.70 &31.86 ±0.47 \\
Random&37.32 ±0.26 &37.14 ±1.30 &33.16 ±0.55 \\
Oracle&43.34 ±1.29 &43.32 ±1.06 &\textbf{43.32 ±0.85} \\
PENS&45.72 ±1.34 &\textbf{43.49 ±0.28} &36.64 ±0.58 \\
\midrule
PANMLoss&\textbf{47.12 ±1.30} &\textbf{45.78 ±1.85} &41.43 ±1.83\\
PANMGrad&\textbf{45.84 ±1.92} &42.14 ±1.34 &\textbf{43.99 ±1.26}  \\
\bottomrule
\end{tabular}
\begin{tabular}{c|ccc}
\toprule
\ &\multicolumn{3}{c}{Precision of Neighbors}\\
\midrule 
Methods&Label-swap(2)&Label-swap(4)&Rotation(4)\\
\midrule
PENS&\textbf{100} &70.83±4.17 &59.75±5.08 \\
PANMLoss&\textbf{100} &\textbf{100} &68.19±28.12\\
PANMGrad&\textbf{100} &83.33±28.87 &\textbf{100} \\
\bottomrule
\end{tabular}
\begin{tabular}{c|ccc}
\toprule
\ &\multicolumn{3}{c}{Recall of Neighbors}\\
\midrule 
Methods&Label-swap(2)&Label-swap(4)&Rotation(4)\\
\midrule
PENS&59.18±2.04 &66.67±4.17 &52.78±8.67 \\
PANMLoss&74.15±32.42 &48.61±13.39 &62.50±31.46\\
PANMGrad&\textbf{100} &\textbf{94.44±6.36} &\textbf{98.61±2.41} \\
\bottomrule
\end{tabular}
\label{dffnoniid}
\end{table}

\begin{figure*}[!t]
\centering
\includegraphics[width=2\columnwidth]{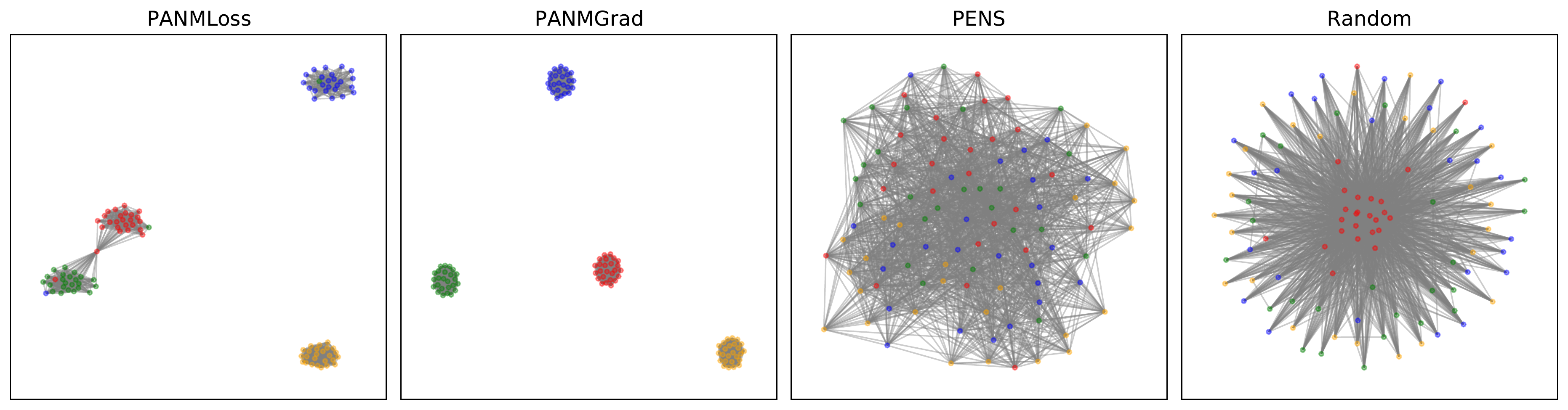}
\caption{Neighbor topologies in the second stage. Four clusters with rotations, CIFAR10, $n=100,l=10,n=5$, trainset size is 200. Each color denotes a cluster.}
\label{topology}
\end{figure*}

For precision (the fraction of true neighbors in the neighbor bag) and recall (the fraction of estimated true neighbors among all true neighbors) of neighbor estimation in the second stage, our methods outperform \texttt{PENS}. It indicates our EM-based \texttt{NAEM} method is effective in enabling clients to match most of the true neighbors. We also demonstrate the neighbor topology in the second stage in Figure \ref{topology}. In \texttt{PANM}, clients evolve to form the four-cluster structure without prior knowledge of cluster identities, whereas \texttt{PENS} and \texttt{Random} construct disordered topologies.

\begin{figure}[t]
\centering
\includegraphics[width=1\columnwidth]{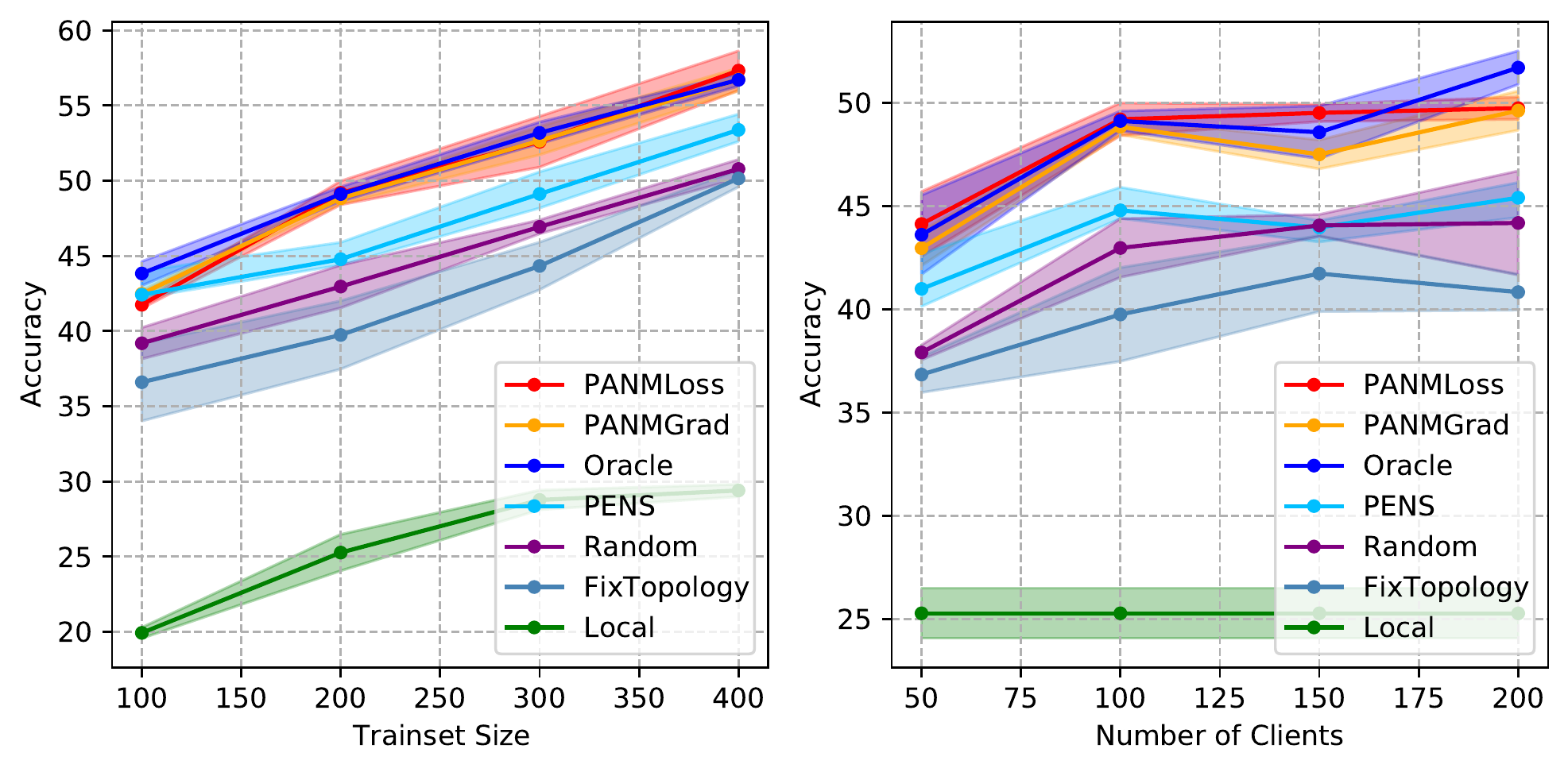} 
\caption{Left: Accuracies when changing trainset size; $n=100,l=10,k=5$, two rotations \{0°,180°\} for all. Right: Accuracies when changing number of clients, $l=10,k=5$, trainset size is 200, two rotations \{0°,180°\} for all.}
\label{figtrainsize}
\end{figure}

\begin{table*}[!t] 
\normalsize
\caption{Impact of $k$ and $l$. CIFAR10 with two rotations \{0°,180°\}, $n=100$, trainset size is 100 for all settings.}
\centering
\begin{tabular}{c|cccccc}
\toprule
\ &\multicolumn{6}{c}{$l, k$}\\
\midrule 
Methods&10,5 &10,3 &20,10 &20,5 &30,15 &30,10\\
\midrule
Local&19.93 ±0.36 &19.93 ±0.36 &19.93 ±0.36&19.93 ±0.36 &19.93 ±0.36 &19.93 ±0.36\\
\midrule 
FixTopology&36.60 ±2.58 &31.54 ±0.18 &38.31 ±1.45 &34.94 ±0.84 &38.88 ±0.52 &38.77 ±3.80\\
Random&39.19 ±1.04 &38.54 ±2.24 &40.82 ±1.09 &37.34 ±0.80 &42.22 ±2.25 &39.91 ±0.60\\
Oracle&\textbf{43.83 ±0.80} &\textbf{42.28 ±1.25} &\textbf{43.39 ±2.80} &\textbf{43.57 ±0.93} &\textbf{44.79 ±2.36} &\textbf{44.16 ±1.32}\\
PENS&42.42 ±1.58 &\textbf{40.04 ±0.90} &42.57 ±1.74 &41.92 ±1.28 &44.12 ±0.22 &42.80 ±0.98\\
\midrule
PANMLoss&41.75 ±0.24 &39.55 ±1.78 &\textbf{44.02 ±0.59} &\textbf{43.64 ±1.54} &\textbf{46.82 ±0.41} &\textbf{46.19 ±0.84}\\
PANMGrad&\textbf{42.48 ±0.21} &39.53 ±2.99 &38.24 ±4.26 &41.94 ±3.15 &41.16 ±3.60 &42.23 ±1.02\\
\bottomrule
\end{tabular}
\label{lk}
\end{table*}

\subsection{Impact of \textit{l} and \textit{k}} \label{subsec:lk}
$l$ is the size of neighbor candidate list, and $k$ is the size of the aggregation neighbor list. The choices of $l$ and $k$ depend on communication budgets in the system, and they determine the network connectivity. Larger $l$ and $k$ will result in denser network connectivity and bring more communication costs. Besides, the ratio of $l$ and $k$ is also crucial (especially for \texttt{PENS}), and it decides the purity of neighbors as we have inferred in Theorem \ref{thm1} and Corollary \ref{corollary1}. We conduct experiments under different network setups by changing $l, k$ as demonstrated in Table \ref{lk}. 

We notice if $l$ and $\frac{k}{l}$ are small (for example, $l=10,k=3$), \texttt{PENS} performs well, but in other settings, \texttt{PENS} has poor results. As we have discussed in Section \ref{subsect:theory}, if $l, k$ are not set appropriately, \texttt{PENS} is prone to be noisy in neighbor matching. 

For \texttt{Oracle}, the communication only relies on $k$. As $k$ increases, the network connection is denser, and the performance of \texttt{Oracle} increases. We explain that if the neighbor bag is pure and the number of aggregation neighbors increases, it is better for clients to reach partial consensus within clusters, and the model weights are more likely to be similar. It is worth mentioning that \texttt{PANMLoss} is very robust under various network setups, surpassing \texttt{Oracle} in most settings. \texttt{PANMLoss} benefits a lot when $k$ increases. 

\begin{table}[!ht] \small
\caption{Comparison with centralized FL. $n=100,l=10,k=5$, trainset size is 200. Best performances in the centralized and decentralized are in bold. Clusters are generated by rotations: 2 for \{0°,180°\}, 4 for \{0°,90°,180°,270°\}.}
\centering
\begin{tabular}{c|ccc}
\toprule
Methods&CIFAR10(4)&FMNIST(4)&FMNIST(2)\\
\midrule 
Local&25.27 ±1.21 &76.24 ±0.22 &76.24 ±0.22\\
\midrule 
FedAvg&37.03 ±0.74&83.54 ±0.08 &86.86 ±0.16 \\
IFCA($c$=2)&40.64 ±2.18 &86.19 ±0.04 &\textbf{88.06 ±0.20} \\
IFCA($c$=3)&41.05 ±1.09 &\textbf{86.78 ±0.36} &/ \\
IFCA($c$=4)&\textbf{43.65 ±0.77} &86.50 ±0.07 &/ \\
\midrule
Oracle&43.32 ±0.85 &85.45 ±0.38 &87.01 ±0.26 \\
PENS&36.64 ±0.58 &84.68 ±0.27 &86.82 ±0.11 \\
\midrule
PANMLoss&41.43 ±1.83 &\textbf{86.09 ±0.31} &\textbf{87.33 ±0.17} \\
PANMGrad&\textbf{43.99 ±1.26} &85.64 ±0.25 &86.88 ±0.34 \\
\bottomrule
\end{tabular}
\label{ifcatable}
\end{table}

\begin{table}[!ht]
\caption{Performances under low communication budgets, where $n = 100$, the trainset size is 200 and $\Omega = 100\omega$ with $\omega$ being the single model size. We use CIFAR-10 with 4 rotations. For PANM, we set $l = 4,~\tau=10$.}
\centering
\begin{tabular}{c|ccc} 
\toprule
Methods &Comm. costs&Max. req. band.&Test acc.\\
\midrule
FedAvg&600$\Omega$ &1$\Omega$ &37.03 ±0.74 \\
IFCA ($c$=2)&900$\Omega$ &2$\Omega$ &40.64 ±2.18 \\
IFCA ($c$=3)&1200$\Omega$ &3$\Omega$ &41.05 ±1.09 \\
IFCA ($c$=4)&1500$\Omega$ &4$\Omega$ &\textbf{43.65 ±0.77} \\
\midrule
PANMLoss ($k$=2)&1118$\Omega$ &0.06$\Omega$ &41.36±0.64 \\
PANMGrad ($k$=2)&1118$\Omega$ &0.06$\Omega$ &42.78±1.68 \\
PANMLoss ($k$=3)&1397$\Omega$ &0.07$\Omega$ &43.30±1.32 \\
PANMGrad ($k$=3)&1397$\Omega$ &0.07$\Omega$ &\textbf{43.34±0.85} \\
\bottomrule
\end{tabular}
\label{tablowcom}
\end{table}

\subsection{Impact of Trainset size and Number of Clients}
In the left figure of Figure \ref{figtrainsize}, we compare the methods by varying the size of the local trainsets, and in the right figure of Figure \ref{figtrainsize}, we show the results of changing the number of clients. We see that \texttt{PANM} consistently outperforms all baselines (\texttt{Local}, \texttt{PENS}, \texttt{Random}, and \texttt{FixTopology}), and it has comparable performance with \texttt{Oracle}.

\subsection{Comparison with Centralized Clustered FL}
P2P FL takes advantage of bandwidth and reliability, as we addressed in Section \ref{intro}. Besides, as for clustered FL, our P2P solution is more robust and can exploit the latent cluster structure in a self-evolved manner without assuming the number of clusters. We compare \texttt{IFCA} \cite{DBLP:conf/nips/GhoshCYR20}, the state-of-the-art centralized clustered FL and centralized \texttt{FedAvg} \cite{DBLP:conf/aistats/McMahanMRHA17} with decentralized P2P methods, as shown in Table \ref{ifcatable} and (c) of Figure \ref{acclosscifar}.
In typical centralized FL algorithms, the central server randomly samples $l$ clients for aggregation, but we notice this will result in bad convergence for \texttt{IFCA}. In our implementations, in a scenario where there are 100 clients with 4 clusters, and the central server samples 10 clients in each round (where $l/n=0.1$), \texttt{IFCA} has poor convergence of estimations. As a result, we have to set full aggregation participation of clients in \texttt{IFCA} (where $l/n=1.0$), but we remind this will cause large communication burdens, and it is unfair to the P2P setting (where we set $n=100, l=10, l/n=0.1$). Even if, in Table \ref{ifcatable}, our P2P method PANM also achieves proportionate performances as a contrast to \texttt{IFCA}. What's more, \texttt{IFCA} requires the assumption on the number of clusters ($c$), and we find that if $c$ is set inappropriately, the performance will be poor. According to Table \ref{ifcatable}, in CIFAR10 with 4 rotations setting, if set $c=2$, the accuracy of \texttt{IFCA} is 40.64\% while our \texttt{PANMLoss} and \texttt{PANMGrad} reach 41.43\% and 43.99\%. Moreover, from the learning curves in Figure (c) of Figure \ref{acclosscifar}, it is apparent that our PANM keeps a more steady and robust learning process while there are some disturbances in the curves of centralized counterparts.

\noindent\textbf{Comparison under Sparse Network.} Generally, if $l$ and $k$ are large, the overall communication costs of P2P FL are larger than those of centralized FL methods. In Table \ref{tablowcom}, we conduct experiments where \texttt{PANM} will have comparable communication costs as the centralized methods. This setting depicts an extremely sparse network for \texttt{PANM}. The results show that \texttt{PANM} can have intimate performances to \texttt{IFCA} when the communication costs are similar, but \texttt{PANM} requires much lower bandwidth. We also find that the consequence of inappropriately estimating the number of clusters will be severe for \texttt{IFCA} (43.65\% when $c=4$; 41.05\% when $c=3$). On the contrary, \texttt{PANM} is flexible enough to explore the cluster structure under any hyperparameters, and the influence of communication budgets is subtle. Therefore, we reckon \texttt{PANM} is more robust and effective than the centralized \texttt{IFCA}.  

\subsection{Results under Real-world Clustered Heterogeneity}
We validate the algorithms on Digit-five benchmark to see how they perform under real-world clustered heterogeneity, and the results are demonstrated in Table \ref{digit-5}.
\begin{table}[!t] 
\caption{Results on Digit-five, $l=10, k=5$, and trainset size is 200.}
\centering
\begin{tabular}{c|cc}
\toprule
&\multicolumn{2}{c}{Number of Clients ($n$)} \\
\midrule
Methods &200 &400\\
\midrule
Local &16.58±0.14 &18.38±0.92 \\
\midrule
FedAvg &80.94±7.66	&71.92±15.8 \\
IFCA ($c$=2) &50.23±1.04 &66.34±23.59 \\
IFCA ($c$=3) &\textbf{91.34±0.44} &52.56±8.82 \\
IFCA ($c$=4) &64.94±28.33 &78.73±13.09 \\
\midrule
FixTopology &73.44±25.14 &39.29±5.13 \\
Random &65.9±19.46 &41.82±9.04 \\
Oracle &\textbf{91.48±0.44} &\textbf{83.05±12.41} \\
PENS &85.57±1.05 &\textbf{86.52±1.03} \\
\midrule
PANMLoss &\textbf{88.6±1.63}	&\textbf{87.04±0.31} \\
PANMGrad &\textbf{89.66±0.51} &\textbf{90.24±1.51} \\
\bottomrule
\end{tabular}
\label{digit-5}
\end{table}
It shows \texttt{PANM} is robust and effective that it has high mean and low variance in accuracy even in the real-world clustered heterogeneity. When the number of clients is 400, \texttt{PANM} has even superior performance over \texttt{Oracle}. 

Interestingly, we observe that even though the Digit-five dataset has four domains, the best assumed number of clusters is not necessarily four. For \texttt{IFCA}, when $n=200$, the best $c$ is 3, and when $n=400$, the best $c$ is 4. It reveals that in real-world data, the clustered structure is not absolute. When the number of clients changes that clients come in and out, the cluster relationship among them also varies. Therefore, it is not flexible of \texttt{IFCA} to assume the cluster number that inappropriate $c$ will cause poor and unstable performances. However, our method can dynamically form the clustered topology, which is more robust and effective.

\section{Discussion on the Applicability of \texttt{PANM}} \label{sect:discussion}
\textbf{Choice between \texttt{PANMLoss} and \texttt{PANMGrad}.} For \texttt{PANMLoss} and \texttt{PANMGrad}, the metrics require different computation resources. As we have discussed in Section \ref{subsect:panm}, generally, \texttt{PANMGrad} is more computationally efficient since it only requires several inner product calculations on sparse vectors (gradients), while \texttt{PANMLoss} needs inference on local datasets. Hence, under limited computation resources, \texttt{PANMGrad} is preferable. 

In addition, in conjunction with the results in Table \ref{lk} and Table \ref{tablowcom}, we found \texttt{PANMGrad} has better performance under sparse communication while \texttt{PANMLoss} benefits more in dense connection. Therefore, under limited communication resources, \texttt{PANMGrad} is preferable. And when the computation and communication resources are sufficient, \texttt{PANMLoss} is the better choice with better performance.

\noindent\textbf{Client Accessibility.} We have mentioned in Subsection \ref{subsec:lk} that we can choose different $l,k$ by the communication budgets, and we have shown \texttt{PANM} is still effective under low communication budgets. Furthermore, in our implementations, we assume fully connected communication accessibility, which means that each client can communicate with any other client in the system. The fully connected accessibility is not often satisfied in realistic scenarios, but we state that \texttt{PANM} is also applicable under limited accessibility. In \texttt{PANM}, we hold loose assumptions about accessibility, that clients can purify their neighbors within the scope of accessible peers. Thus, \texttt{PANM} is flexible in practice.

\section{Conclusion} \label{sect:conclusion}
This paper studies the clustered Non-IID problem in FL under P2P communication and develops \texttt{PANM} that enables clients to match neighbors with similar objectives. \texttt{PANM} is more flexible and effective than the centralized clustered FL methods because it does not require the assumption of the number of clusters. Specifically, in \texttt{PANM}, we present two novel metrics for measuring client similarity based on loss and gradient, respectively. Then, we propose a two-stage algorithm. In the first stage, an effective method based on Monte Carlo is proposed to enable clients to match neighbors with maximally high similarities. Then in the second stage, a method based on Expectation Maximization under the Gaussian Mixture Model assumption of similarities is used for clients to discover more neighbors with similar objectives. We have conducted theoretical analyses of \texttt{PANM} on the probability of neighbor estimation and the error gap to the clustered optimum. We have also implemented extensive experiments under both synthetic and real-world clustered heterogeneity. Theoretical analysis and empirical experiments show that the proposed algorithm is superior to the P2P FL counterparts and achieves better performance than the centralized cluster FL method. \texttt{PANM} is effective even under extremely low communication budgets.

\ifCLASSOPTIONcompsoc
  \section*{Acknowledgments}
\else
  \section*{Acknowledgment}
\fi
This work is supported by the National Key Research and Development Project of China (No. 2021ZD0110400), National Natural Science Foundation of China (No. U19B2042), The University Synergy Innovation Program of Anhui Province (No. GXXT-2021-004), and Key Research Project of Zhejiang Lab (No. 2021KE0AC02).

\ifCLASSOPTIONcaptionsoff
  \newpage
\fi



%\bibliographystyle{IEEEtran}

% \input{ieee.bbl}
\bibliographystyle{IEEEtran}
\bibliography{ieee}


\begin{IEEEbiography}[{\includegraphics[width=1in,height=1.25in,clip,keepaspectratio]{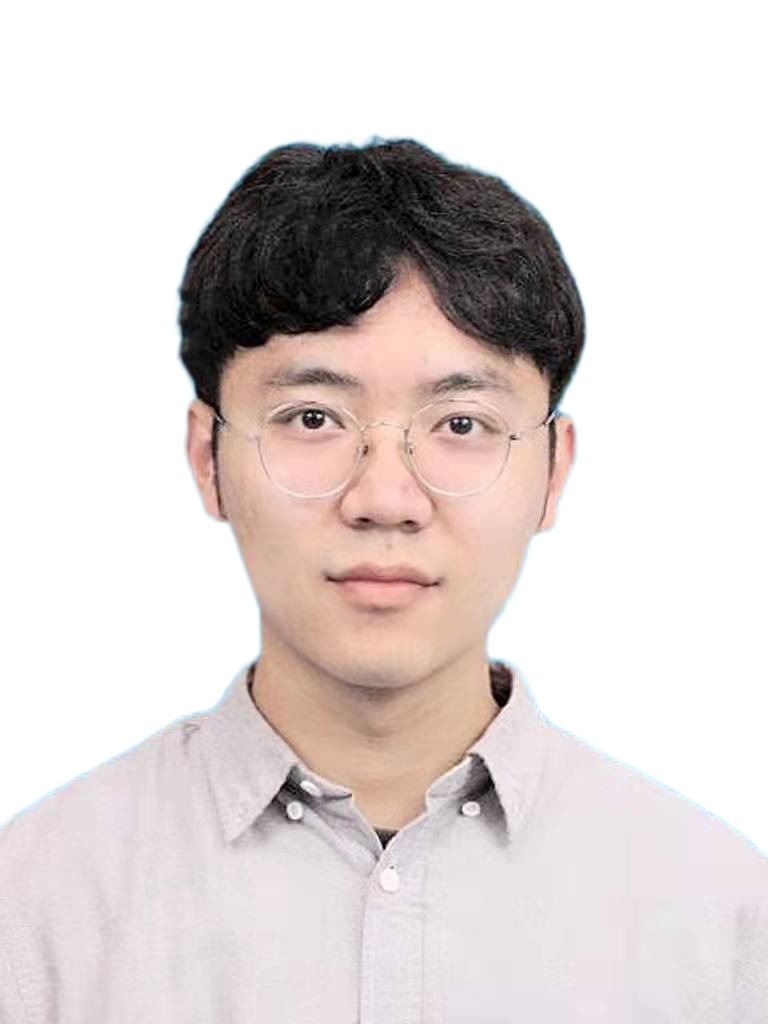}}]{Zexi~Li} received B.S. degree in Agriculture Engineering from Zhejiang University, Hangzhou, China, in 2020. Meanwhile, he also obtained an honorary degree from Chu Kochen Honors College, Zhejiang University. He is currently a Ph.D. student in Computer Science in Zhejiang University. His main research interests include federated learning, optimization, and deep learning.
\end{IEEEbiography}

\begin{IEEEbiography}[{\includegraphics[width=1in,height=1.25in,clip,keepaspectratio]{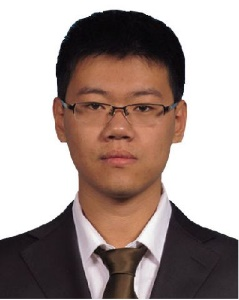}}]{Jiaxun~Lu} (\textit{Member, IEEE}) received the B.S. and Ph.D. degrees from the Department of Electronic Engineering, Tsinghua University, Beijing, China, in 2014 and 2019, respectively. From September 2017 to September 2018, he was a Visiting Student with Coordinate Science Laboratory, University of Illinois at Urbana-Champaign, Champaign, IL, USA. Currently, He is a senior researcher with Noah’s Ark Lab, Huawei Technologies. His research interests include statistical learning, information theory, and mobile wireless communication.
\end{IEEEbiography}

\begin{IEEEbiography}[{\includegraphics[width=1in,height=1.25in,clip,keepaspectratio]{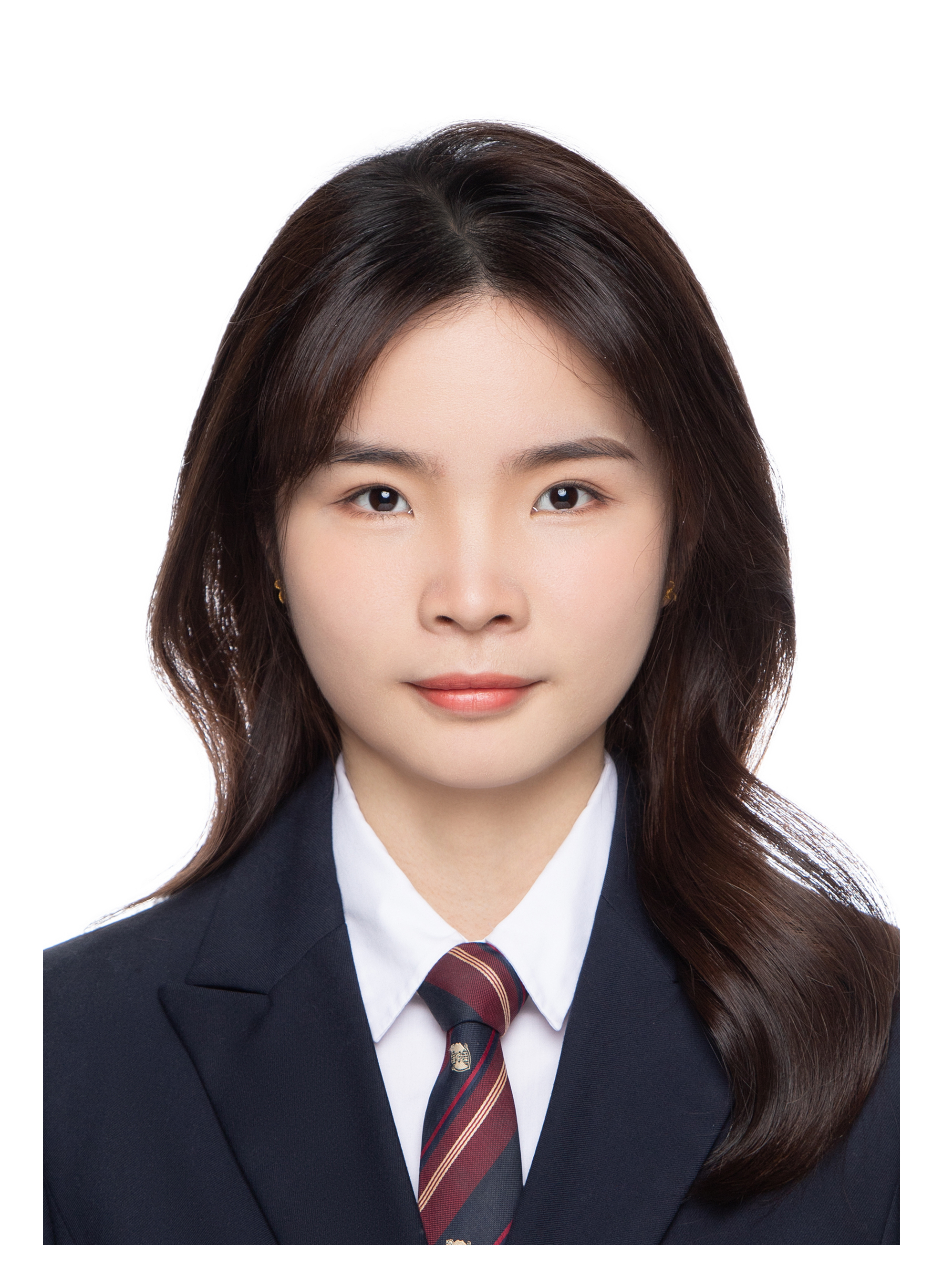}}]{Shuang~Luo} received the B.Eng. degree from the School of Data and Computer Science, Sun Yat-sen University, China, in 2019. She is currently pursuing the Ph.D. degree in the College of Public Affairs, Zhejiang University. Her current research interests include deep learning, reinforcement learning and distributed artificial intelligence.
\end{IEEEbiography}

\begin{IEEEbiography}[{\includegraphics[width=1in,height=1.25in,clip,keepaspectratio]{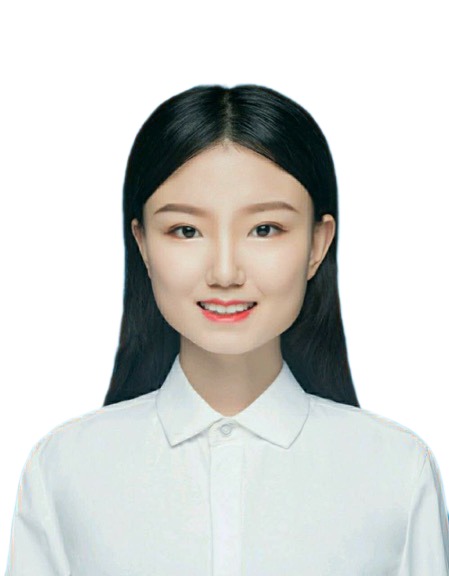}}]{Didi~Zhu} received the B.S. degree in Computer Science and Technology from Beijing University of Chemical Technology, Beijing, China, in 2020. She is currently a Ph.D. student of Zhejiang University. Her main research interests include federated learning and domain adaptation.
\end{IEEEbiography}

\begin{IEEEbiography}[{\includegraphics[width=1in,height=1.25in,clip,keepaspectratio]{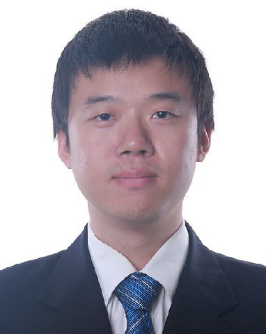}}]{Yunfeng~Shao} (\textit{Member, IEEE}) received the B.S. and Ph.D. degrees in Electronic Engineering from Shanghai Jiao Tong University and University of Chinese Academy of Sciences China, in 2009 and 2014, respectively. He is currently an expert in Huawei Noah’s Ark Lab. His research interests include machine learning with privacy protection, federated learning, transfer learning and their applications in Telecommunication network.
\end{IEEEbiography}

\begin{IEEEbiography}[{\includegraphics[width=1in,height=1.25in,clip,keepaspectratio]{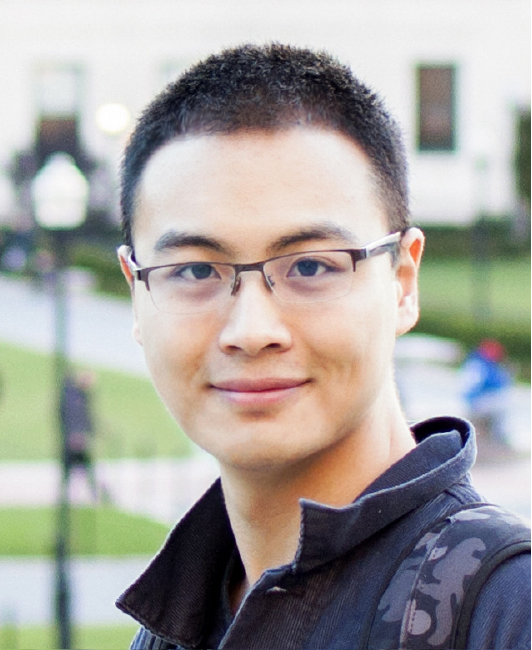}}]{Yinchuan~Li} (\textit{Member, IEEE}) was born in Bozhou, Anhui, China, in 1994. He received the B.S. and Ph.D. degrees in electronic engineering from the Beijing Institute of Technology (BIT), Beijing, China, in 2015 and 2020, respectively. From November 2017 to November 2019, he was a Visiting Scholar with the Department of Electrical Engineering, Columbia University, New York, NY, USA. From February 2020 to August 2020, he was a senior technical consultant in Sant\'e Ventures, Austin, TX, USA. He now works at Noah's Ark Lab, Huawei Technologies, Beijing, China, as an AI researcher. His current research interests include machine learning, deep learning, reinforcement learning, and sparse signal processing.

Dr. Li received the Excellent Paper Award at the 2019 IEEE International Conference on Signal, Information and Data Processing.
\end{IEEEbiography}

\begin{IEEEbiography}[{\includegraphics[width=1in,height=1.25in,clip,keepaspectratio]{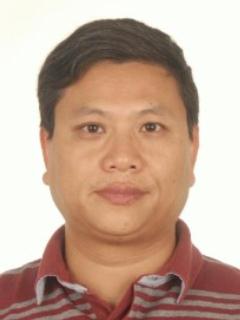}}]{Zhimeng~Zhang} received Ph.D. degree from the College of Computer Science and Technology, Zhejiang University in 2004. He completed postdoctoral career in system control from Zhejiang University in 2011. Since 2004, he has been teaching software development courses for master students in software engineering. From 2006, he has instructed more than 200 master students in the School of Software Technology, Zhejiang University.
\end{IEEEbiography}

\begin{IEEEbiography}[{\includegraphics[width=1in,height=1.25in,clip,keepaspectratio]{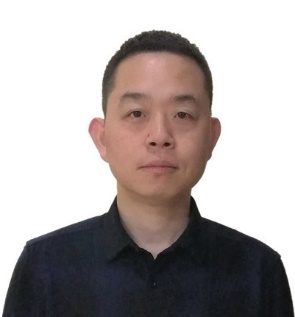}}]{Yongheng~Wang} received his PhD degree in computer science and technology from National University of Defense Technology, Changsha, China, in 2006. Currently he is a research specialist at the research center of Big data intelligence, Zhejiang Lab. His research interest covers Big data analysis, machine learning, computer simulation and intelligent decision making.
\end{IEEEbiography}

\begin{IEEEbiography}[{\includegraphics[width=1in,height=1.25in,clip,keepaspectratio]{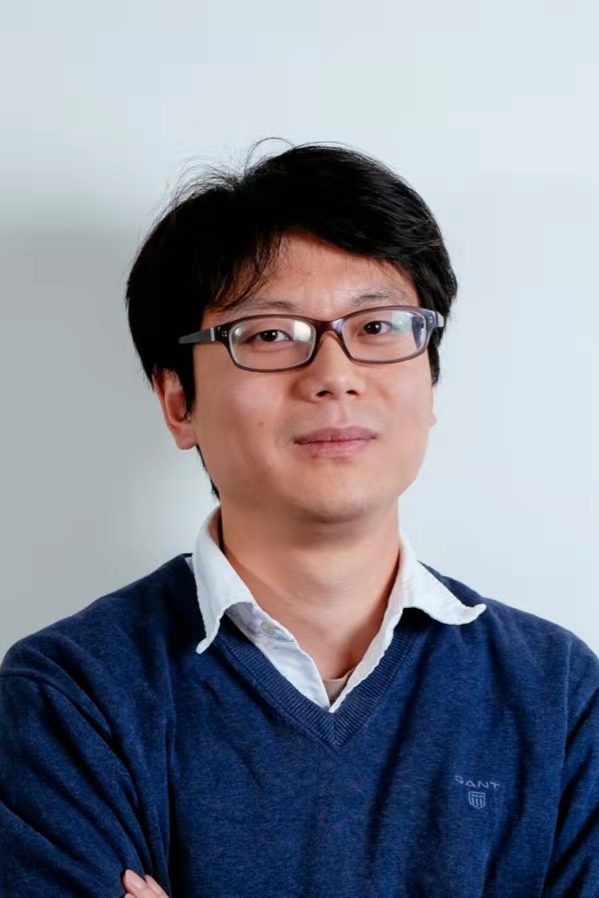}}]{Chao~Wu} is a tenure-track associate professor at the School of Public Affairs, Zhejiang University and the director of Computational Social Science Research Center in Zhejiang University. He is also an honorary Research Fellow at The Department of Computer Science, Imperial College London. His research interests include federated learning and distributed machine learning, data privacy protection and data pricing, and computational social sciences. He has published more than 60 papers in international conferences and journals, and presided over many scientific research projects including the key projects of the National Natural Science Foundation of China.  
\end{IEEEbiography}


% % insert where needed to balance the two columns on the last page with
% %\newpage



%\vfill

\appendices
\section{Proof of Theorem 1 and Corollary 1}
\label{apdx:section1}
\subsection{Proof of Theorem 1}

Under Assumption 2, the similarity metrics are effective enough that all the true neighbors will have higher similarities than the false neighbors. Therefore if there are at least $k$ true neighbors in the $l$ neighbor candidates, the top $k$ peers with maximum similarities must be true. As a result, we can formulate the problem of all the top $k$ neighbors being true into the \textbf{\textit{``Ball Selection Problem''}}, as follows.

\noindent \textit{\textbf{Ball Selection Problem:} There are $n-1$ balls in the box and $a-1$ of them are white balls. Select $l$ balls at one time without replacement. For the selected $l$ balls, solve: \\ (1) the probability that at least $k$ balls are white balls. \\ Conducting the selection for the first time and note down the number of white balls as $s_{1}$.  Conducting the selection for the second time and note down the number of white balls as $s_{2}$. \dots. Conducting the selection for the $t$-th time and note down the number of white balls as $s_{t}$. Solve: \\ (2) the probability that $\sum\limits_{c=1}^{t} s_{c} \geq k$.}

Question (1) in the ``Ball Selection Problem'' is equivalent to the problem that for client $i$, randomly sampling $l$ clients, solve the probability that at least $k$ peers are true. Question (2) in ``Ball Selection Problem'' is equivalent to the problem that for client $i$, conducting \texttt{NSMC} strategy for $t$ rounds, solve the probability that all the $k$ neighbors in round $t$ are true. 

We further calculate the probabilities in Questions (1) and (2). We define the probability that select $l$ balls and $x$ of them are white balls as a function ${\rm G}(x)$ and the probability that select $l$ balls and at least $x$ of them are white balls as function ${\rm R}(x)$. 

First, we give the calculation of ${\rm G}(x)$. All possible cases are: from all $n-1$ balls to select $l$ balls, as $C_{n-1}^{l}$. The cases satisfying our condition are: from $a-1$ white balls to select $x$ balls while from other $n-a$ balls to select $l-x$ balls, as $C_{a-1}^{x}\cdot C_{n-a}^{l-x}$. The equation is as follows.

\begin{equation} \small
\begin{split}
{\rm G}(x) =&\frac{C_{a-1}^{x}\cdot C_{n-a}^{l-x}}{C_{n-1}^{l}} \\
=&\bigg[\frac{(a-1)!}{x!(a-x-1)!}\cdot \frac{(n-a)!}{(l-x)!(n-a-l-x)!}\bigg] \bigg/\\
&\bigg[\frac{(n-1)!}{l!(n-l-1)!}\bigg]\\
=&\frac{l!(a-1)!(n-a)!(n-l-1)!}{x!(n-1)!(l-x)!(a-x-1)!(n-a-l+x)!}.
\end{split}
\end{equation}

Then, we give the calculation of ${\rm R}(x)$. All possible cases are: from all $n-1$ balls to select $l$ balls, as $C_{n-1}^{l}$. The number of white balls satisfying our condition ranges from $x$ to $l$, so the cases satisfying our condition are: $x$ white balls and $l-x$ other balls, $x+1$ white balls and $l-x-1$ other balls, \dots, $l-1$ white balls and $1$ other balls, $l$ white balls and $0$ other ball. Thus the equation is as follows.

\begin{equation} \small
\begin{split}
{\rm R}(x) =& \big( C_{a-1}^{x}\cdot C_{n-a}^{l-x} + C_{a-1}^{x+1}\cdot C_{n-a}^{l-x-1} + \dots + C_{a-1}^{l-1}\cdot C_{n-a}^{1}+\\
&C_{a-1}^{l}\cdot C_{n-a}^{0} \big)\big/C_{n-1}^{l}\\
=&\Bigg[\sum\limits_{s=0}^{l-x}\frac{(a-1)!}{(l-s)!(a-l+s-1)!}\cdot \frac{(n-a)!}{s!(n-a-s)!}\Bigg] \bigg/\\
&\bigg[\frac{(n-1)!}{l!(n-l-1)!}\bigg]\\
=& {\frac{l!(n-l-1)!}{(n-1)!}\sum\limits_{s=0}^{l-x} \frac{(a-1)!(n-a)!}{s!(l-s)!(n-a-s)!(a-l+s-1)!}}\\.
\end{split}
\end{equation}

Obviously, for Question (1), the probability is ${\rm P}^{1}(k) = {\rm R}(k)$. Then, we consider the scenario where $t=2$ in Question (2). All the satisfying cases are: none of the white balls is selected in the first time and at least $k$ white balls are selected in the second time, 1 white ball is selected in the first time and at least $k-1$ white balls are selected in the second time,\dots, $k-1$ white balls are selected in the first time and at least $1$ white balls are selected in the second time, at least $k$ white balls are selected in the first time and it does not matter how many white balls are selected in the second time. Therefore, the equation is as follows. 

\begin{equation} \small
\begin{split}
{\rm P}^{2}(k) =& {\rm G}(0)\cdot {\rm R}(k) + {\rm G}(1)\cdot {\rm R}(k-1)+ \dots + {\rm G}(k-1)\cdot \\
&{\rm R}(1) + {\rm R}(k).
\end{split}
\label{equ:p2}
\end{equation}

Similar to Equation (\ref{equ:p2}), we can infer the satisfying cases when $t=t$: none of the white balls is selected in the first time and at least $k$ white balls are selected in the remaining $t-1$ times, 1 white ball is selected in the first time and at least $k-1$ white balls are selected in the remaining $t-1$ times,\dots, $k-1$ white balls are selected in the first time and at least $1$ white balls are selected in the remaining $t-1$ times, at least $k$ white balls are selected in the first time and it does not matter how many white balls are selected in the remaining $t-1$ times. The equation is as follows.

\begin{equation} \small
\begin{split}
{\rm P}^{t}(k) =& {\rm G}(0)\cdot {\rm P}^{t-1}(k) + {\rm G}(1)\cdot {\rm P}^{t-1}(k-1)+ \dots + {\rm G}(k-1)\cdot \\
&{\rm P}^{t-1}(1) + {\rm R}(k)\\
=& \sum\limits^{k-1}_{m=0} {\rm G}(m){\rm P}^{t-1}(k-m) + {\rm R}(k)\\
=& {\rm G}(k)\ast{\rm P}^{t-1}(k) + {\rm R}(k),
\end{split}
\label{equ:pt}
\end{equation}
where ${\rm G}(k)\ast{\rm P}^{t-1}(k) = \sum\limits^{k-1}_{m=0} {\rm G}(m){\rm P}^{t-1}(k-m)$.
\\
\\
\indent For the first stage of \texttt{PENS}, neighbor selection in each round is independent, so the probability remains unchanged in all rounds, we have

\begin{equation}
    {\rm P}^{t}(k) \equiv {\rm R}(k).
\end{equation}

\subsection{Proof of Corollary 1}

Function ${\rm Q}(t)$ is defined as ${\rm Q}(t) = {\rm P}^{t}(k), t\in \{1,\dots,T\}$. To prove Corollary 1, we need to prove ${\rm Q}(t) - {\rm Q}(t-1) > 0, t\in \{2,\dots,T\}$. We use Mathematical Induction method to prove the corollary.

\noindent\textbf{Step 1:} ${\rm Q}(t) - {\rm Q}(t-1) > 0$ is satisfied when $t=2$, because 
\begin{equation} 
\begin{split}
    {\rm Q}(2)-{\rm Q}(1)=&  {\rm P}^{2}(k) -  {\rm P}^{1}(k)\\
    =& {\rm G}(k)\ast {\rm P}^{1}(k) + {\rm R}(k) - {\rm R}(k)\\
    =& {\rm G}(k)\ast {\rm P}^{1}(k) > 0.
\end{split}
\end{equation}
% {\rm P}^{2}(k) &= {\rm G}(k)\ast {\rm P}^{1}(k) + {\rm R}(k)\\
% {\rm P}^{1}(k) &= {\rm R}(k).
\noindent\textbf{Step 2:} Assume ${\rm Q}(t) - {\rm Q}(t-1) > 0$ when $t=t$, we now prove ${\rm Q}(t+1) - {\rm Q}(t) > 0$ when $t=t+1$. 

\begin{equation} 
\begin{split}
    &{\rm Q}(t+1)-{\rm Q}(t)\\
    =& {\rm P}^{t+1}(k)-{\rm P}^{t}(k)\\
    =& {\rm G}(k)\ast {\rm P}^{t}(k) + {\rm R}(k) - ({\rm G}(k)\ast {\rm P}^{t-1}(k) + {\rm R}(k))\\
    =& {\rm G}(k)\ast {\rm P}^{t}(k) - {\rm G}(k)\ast {\rm P}^{t-1}(k)\\
    =& \sum\limits^{k-1}_{m=0} {\rm G}(m){\rm P}^{t}(k-m)-\sum\limits^{k-1}_{m=0} {\rm G}(m){\rm P}^{t-1}(k-m)\\
    =& \sum\limits^{k-1}_{m=0} {\rm G}(m)[{\rm P}^{t}(k-m)-{\rm P}^{t-1}(k-m)].
\end{split}
\label{equa:coro1}
\end{equation}
We have ${\rm Q}(t) - {\rm Q}(t-1) > 0$, further
\begin{equation} 
    {\rm Q}(t)-{\rm Q}(t-1)= {\rm P}^{t}(k)-{\rm P}^{t-1}(k)>0.
\end{equation}
Therefore,
\begin{equation} 
{\rm P}^{t}(k-m)-{\rm P}^{t-1}(k-m)>0,\ k>m.
\label{equa:coro2}
\end{equation}
Together with Equation (\ref{equa:coro1}) and (\ref{equa:coro2}), we can infer 
\begin{equation}
    {\rm Q}(t)-{\rm Q}(t-1) > 0.
\end{equation}

\noindent\textbf{Step 3:} Based on Step 1 and 2, we can draw the conclusion that 
\begin{equation}
    {\rm Q}(t) - {\rm Q}(t-1) > 0,\ \forall t\in \{2,\dots,T\}
\end{equation}
Then, the proof is completed.

\section{Proof of Theorem 2 and Theorem 3}
\label{apdx:section2}

\subsection{Proof of Theorem 2}
In a P2P FL system, each client communicates with his neighbors and averages the models received from the neighbors. For algorithms like \texttt{PANM} and \texttt{PENS}, the communicated neighbors change in each round. This system is complex because your model's performance is not only affected by your neighbors (first-order neighbors) but also your neighbors' neighbors (second-order neighbors), $\cdots$. The changing neighbor relationships and the complexity of information flow make the convergence of each client intractable. 

Thus, to make this problem tractable, we only consider the first-order neighbors' impacts and analyze from client perspective. For client $i, i \in [n]$, we view it as the central server and only consider the neighbor communication and model averaging process (without the local training process). And for other clients, we only consider the local training process and assume their local models are downloaded from client $i$ in each round, ignoring their gossip communications. This simplification means that for a given client, we only consider the impact of its first-order neighbors, and the convergence and performance rely on selecting these neighbors. It is reasonable in the theoretical analysis since the first-order neighbors have the most dominant impacts, and when the orders of neighbors are higher, the impacts are much weaker.

Without loss of generality, we analyze client 1, which belongs to cluster 1, and the analysis is general to all clients. Recall that the optimal model of cluster 1 is denoted as $\mathbf{w}_1^*$, for a certain round, client 1's model is $\mathbf{w}_1$, and its next-round model is $\mathbf{w}_1^+$. For simplicity, in the analysis, we omit the superscript $t$, which indicates the communication round. We consider the error gap between the optimum $\mathbf{w}_1^*$ and the next-round model $\mathbf{w}_1^+$.
\begin{footnotesize}
\begin{equation} 
    \Vert \mathbf{w}_{1}^{+} - \mathbf{w}_{1}^{*} \Vert = \Vert \underbrace{\mathbf{w}_1 - \mathbf{w}_1^* - \frac{\eta}{k}\sum_{i\in {\rm N}_1 \cap {\rm N}_1^*} \nabla F_i(\mathbf{w}_1)}_{T_1} - \underbrace{\frac{\eta}{k}\sum_{i\in {\rm N}_1 \cap \overline{{\rm N}_1^*}} \nabla F_i(\mathbf{w}_1)}_{T_2} \Vert,
    \nonumber
\end{equation}
\end{footnotesize}
where $\eta$ is the learning rate and $k~(k = |{\rm N}_1|)$ is the number of aggregation neighbors. We use the triangle inequality and obtain
\begin{equation} \label{T1_T2}
    \Vert \mathbf{w}_{1}^{+} - \mathbf{w}_{1}^{*} \Vert \leq \Vert T_1 \Vert + \Vert T_2 \Vert.
\end{equation}
In Equation \ref{T1_T2}, $\Vert T_1 \Vert$ is the term including the gradients of true neighbors and $\Vert T_2 \Vert$ is the term about the gradients of false neighbors. We bound $\Vert T_1 \Vert$ and $\Vert T_2 \Vert$ respectively.

\noindent \textbf{Bound $\Vert T_1 \Vert$} Recall that $\epsilon = \frac{| {\rm N}_{1} \cap \overline{{\rm N}_{1}^{*}} |}{| {\rm N}_{1} |}$.
\begin{small}
\begin{equation}
\begin{split}
      T_1 =& \underbrace{\mathbf{w}_{1} - \mathbf{w}_{1}^{*} - \eta(1-\epsilon)\nabla F^1(\mathbf{w}_{1})}_{T_{11}} \\
      &+\underbrace{\eta(1-\epsilon)\Big(\nabla F^1(\mathbf{w}_{1}) - \frac{1}{k(1-\epsilon)}\sum_{i\in {\rm N}_1 \cap {\rm N}_1^*}\nabla F_i(\mathbf{w}_1) \Big)}_{T_{12}}
\end{split}
\nonumber
\end{equation}
\end{small}

We assume $\eta \leq \frac{1}{L}$, thus $\eta(1-\epsilon)\leq\frac{1}{L}$. According to standard analysis techniques for gradient descent on strongly convex functions, we know that 
\begin{equation}
\begin{split}
        \Vert T_{11} \Vert &= \Vert \mathbf{w}_{1} - \mathbf{w}_{1}^{*} - \eta(1-\epsilon)\nabla F^1(\mathbf{w}_{1}) \Vert \\
        &\leq \left( 1 - \frac{\mu L\eta(1-\epsilon)}{\mu+L} \right)\Vert \mathbf{w}_{1}-\mathbf{w}_{1}^* \Vert.
\end{split}
    \nonumber
\end{equation}
Further, we have $\mathbb{E}\left[ \Vert T_{12} \Vert^2 \right] = \frac{v^2}{dk(1-\epsilon)}$, which implies that $\mathbb{E}\left[ \Vert T_{12} \Vert \right] = \frac{v}{\sqrt{dk(1-\epsilon)}}$.

Therefore, the bound of $\Vert T_1 \Vert$ is 
\begin{equation} \label{bound_T1}
    \Vert T_1 \Vert \leq \left( 1 - \frac{\mu L\eta(1-\epsilon)}{\mu+L} \right)\Vert \mathbf{w}_{1}-\mathbf{w}_{1}^* \Vert + \frac{v}{\sqrt{dk(1-\epsilon)}}.
\end{equation}

\noindent \textbf{Bound $\Vert T_2 \Vert$} We define $T_{2j} \coloneqq \sum_{i\in {\rm N}_1 \cap {\rm N}_j^*}\nabla F_i(\mathbf{w_1}), j \geq 2, j \in [r]$, which refers to the sum gradients of the neighbors which belong to cluster $j$. Thus, we have $T_2 = \frac{\eta}{k}\sum_{j=2}^r T_{2j}$. For $T_{2j}$, we have
\begin{equation}
    T_{2j} = |{\rm N}_1 \cap {\rm N}_j^*|\nabla F^j(\mathbf{w}_1) + \sum_{i \in {\rm N}_1 \cap {\rm N}_j^*}\left(\nabla F_i(\mathbf{w}_1) - \nabla F^j(\mathbf{w}_1)\right).
    \nonumber
\end{equation}
Due to the smoothness of $F^j(\mathbf{w})$, we know that
\begin{equation}
\begin{split}
    \Vert \nabla F^j(\mathbf{w}_1) \Vert &\leq L\Vert \mathbf{w}_1 - \mathbf{w}_j^*\Vert = L\Vert \mathbf{w}_1 - \mathbf{w}_1^* + \mathbf{w}_1^* - \mathbf{w}_j^* \Vert \\
    &\leq L(\Vert \mathbf{w}_1 - \mathbf{w}_1^*\Vert + \Vert\mathbf{w}_1^* - \mathbf{w}_j^* \Vert) \\
    &\leq L(\Vert \mathbf{w}_1 - \mathbf{w}_1^*\Vert + \Delta).
\end{split}
    \nonumber
\end{equation}
Additionally, according to the bounded variance of gradients, we have 
\begin{equation}
\begin{split}
    &\mathbb{E}\left[ \Vert \sum_{{\rm N}_1 \cap {\rm N}_j^*}\left(\nabla F_i(\mathbf{w}_1) - \nabla F^j(\mathbf{w}_1)\right) \Vert^2 \right] = |{\rm N}_1 \cap {\rm N}_j^*|\frac{v^2}{d},\\
    &\mathbb{E}\left[ \Vert \sum_{{\rm N}_1 \cap {\rm N}_j^*}\left(\nabla F_i(\mathbf{w}_1) - \nabla F^j(\mathbf{w}_1)\right) \Vert \right] = \sqrt{|{\rm N}_1 \cap {\rm N}_j^*|}\frac{v}{\sqrt{d}}.
\end{split}
\nonumber
\end{equation}

Summing up, the bound for $\Vert T_2 \Vert$ is
\begin{equation} \label{bound_T2}
    \Vert T_2 \Vert = \frac{\eta}{k}\Vert \sum_{j=2}^{r} T_{2j} \Vert \leq \eta L\epsilon\Vert \mathbf{w}_1 - \mathbf{w}_j^*  \Vert + \eta L\Delta\epsilon + \frac{v\eta\sqrt{r}}{\sqrt{dk}}\sqrt{\epsilon}.
\end{equation}

Combining Equations \ref{T1_T2}, \ref{bound_T1} and \ref{bound_T2}, we can have a bound for $\Vert T_1 \Vert + \Vert T_2 \Vert$

\begin{equation} \label{bound_T1_T2}
\begin{split}
    \Vert \mathbf{w}_{1}^{+} - \mathbf{w}_{1}^{*} \Vert \leq& \left(1 - \frac{\mu L\eta(1-\epsilon)}{\mu+L} + \eta L\epsilon \right)\Vert \mathbf{w}_1 - \mathbf{w}_j^*  \Vert \\
    &+ \eta L\Delta\epsilon + \frac{v}{\sqrt{dk}}\frac{1}{\sqrt{1-\epsilon}} + \frac{v\eta\sqrt{r}}{\sqrt{dk}}\sqrt{\epsilon}.
\end{split}
\end{equation}
The proof is completed.

\subsection{Proof of Theorem 3}
We set the learning rate $\eta = \frac{1}{L}$, assume $\mu = L$ and define the initial error gap is defined as $\delta_{0} = \Vert \mathbf{w} \Vert$. According to Theorem 2, we can have
\begin{small}
\begin{equation} \label{thm1:equ}
% \begin{split}
    \Vert \mathbf{w}_{1}^{+} - \mathbf{w}_{1}^{*} \Vert \leq (\frac{1+3\epsilon}{2})\Vert \mathbf{w}_{1} - \mathbf{w}_{1}^{*} \Vert + \Delta \epsilon + \frac{v}{\sqrt{dk}}\frac{1}{\sqrt{1-\epsilon}} + \frac{v}{L}\sqrt{\frac{r}{kd}}\sqrt{\epsilon}.
% \end{split}
\nonumber
\end{equation}
\end{small}
For PANM, the initial $\epsilon_0={\rm R}(k)$, thus, for the first round, we have
\begin{footnotesize}
\begin{equation}
\begin{split}
   \Vert \mathbf{w}_{1}^{1} - \mathbf{w}_{1}^{*} \Vert &\leq (\frac{1+3\epsilon_0}{2})\Vert \mathbf{w}_{1}^0 - \mathbf{w}_{1}^{*} \Vert + \Delta \epsilon_0 + \frac{v}{\sqrt{dk}}\frac{1}{\sqrt{1-\epsilon_0}} + \frac{v}{L} \sqrt{\frac{r\epsilon_0}{kd}} \\
   &= (\frac{1+3\epsilon_0}{2})\delta_0 + \Delta \epsilon_0 + \frac{v}{\sqrt{dk}}\frac{1}{\sqrt{1-\epsilon_0}} + \frac{v}{L} \sqrt{\frac{r}{kd}}\sqrt{\epsilon_0}
\end{split}
\end{equation}
\end{footnotesize}
According to Theorem 1, in the first stage of \texttt{PANM}, it will have a decreasing $\epsilon$, and the decreasing speed is very fast. As a result, we only consider $\epsilon_0$ and for $T\geq1$, we assume $\epsilon \rightarrow 0$. When $\epsilon \rightarrow 0$, we have
\begin{equation}
    \Vert \mathbf{w}_{1}^{+} - \mathbf{w}_{1}^{*} \Vert \leq \frac{1}{2}\Vert \mathbf{w}_{1} - \mathbf{w}_{1}^{*} \Vert + \frac{v}{dk}.
    \nonumber
\end{equation}
Summing the error gaps over rounds, we can have the error bound for the first stage of \texttt{PANM}, as
\begin{equation}
\begin{split}
    \Vert \mathbf{w}_{1}^{T} - \mathbf{w}_{1}^{*} \Vert &\leq \frac{1}{2^{T-1}}	{\Bigg[}\frac{1+3\epsilon_{0}}{2}\delta_{0} + \epsilon_{0}\Delta +\frac{v}{\sqrt{dk(1-\epsilon_{0})}} \\
    &+\frac{v}{L}\sqrt{\frac{r\epsilon_{0}}{kd}}\Bigg] + \sum_{t=0}^{T-2}\frac{1}{2^t}\frac{v}{\sqrt{dk}},
\end{split}
\end{equation}
where $\epsilon_{0} = {\rm R}(k) = \small{\frac{l!(n-l-1)!}{(n-1)!}\sum\limits_{s=0}^{l-k} \frac{(a-1)!(n-a)!}{s!(l-s)!(n-a-s)!(a-l+s-1)!}}$.

\end{document}